%% file: main.tex
\documentclass[a4paper,fleqn]{cas-dc}
\usepackage[authoryear]{natbib} 

\usepackage{threeparttable}
\usepackage{xcolor}
\usepackage{graphicx}

\usepackage{subcaption}
\usepackage{tabularx}

\def\tsc#1{\csdef{#1}{\textsc{\lowercase{#1}}\xspace}}
\tsc{WGM}
\tsc{QE}

\newdefinition{rmk}{Remark}
\usepackage{amsmath}
\usepackage{upgreek}
\usepackage{appendix}
\usepackage[pagewise]{lineno}
\usepackage{float}
\usepackage{placeins}
\usepackage{flushend}
\begin{document}
\begin{sloppypar} 

\let\WriteBookmarks\relax
\def\floatpagepagefraction{1}
\def\textpagefraction{.001}
\let\printorcid\relax

\makeatletter
\renewcommand{\fnum@figure}{Fig. \thefigure.\@gobble}
\makeatother

\shorttitle{}    

\shortauthors{}  

\title [mode = title]{AWARE: Adaptive Whole-body Active Rotating Control for Enhanced LiDAR-Inertial Odometry under Human-in-the-Loop Interaction}


\tnotetext[1]{This research was jointly supported by the National Natural Science Foundation Project ( No. 42201477, No.42130105) and Wuhan Natural Science Foundation Project (No. 2025041001010363).} 

\author[1]{Yizhe Zhang}

\ead{yizhezhang0418@whu.edu.cn}
\credit{Conceptualization of this study, Methodology, Software, Data acquisition, Equipment maintenance, Validation, Writing – original draft}
\affiliation[1]{organization={State Key Laboratory of Information Engineering in Surveying, Mapping and Remote Sensing},
            addressline={Wuhan University}, 
            city={Wuhan},
            postcode={430079}, 
            country={China}}

\author[2]{Jianping Li}
\cormark[1]
\ead{jianping.li@ntu.edu.sg}
\credit{Conceptualization of this study, Methodology, Equipment maintenance, Writing review, Funding acquisition}
\affiliation[2]{organization={School of Electrical and Electronic Engineering},
            addressline={Nanyang Technological University}, 
           postcode={639798}, 
            country={Singapore}}

\author[1]{Liangliang Yin}
\ead{2022302141086@whu.edu.cn}
\credit{Data acquisition, Equipment maintenance}

\author[1]{Zhen Dong}
\ead{dongzhenwhu@whu.edu.cn}
\credit{Conceptualization of this study, Methodology, Writing review, Project administration, Funding acquisition}

\author[1]{Bisheng Yang}
\ead{bshyang@whu.edu.cn}
\credit{Conceptualization of this study, Project administration, Funding acquisition}

\cortext[1]{Corresponding author}



\begin{abstract}
Human-in-the-loop (HITL) UAV operation is essential in complex and safety-critical aerial surveying environments, where human operators provide navigation intent while onboard autonomy must maintain accurate and robust state estimation. A key challenge in this setting is that resource-constrained UAV platforms are often limited to narrow-field-of-view LiDAR sensors. In geometrically degenerate or feature-sparse scenes, limited sensing coverage often weakens LiDAR Inertial Odometry (LIO)'s observability, causing drift accumulation, degraded geometric accuracy, and unstable state estimation, which directly compromise safe and effective HITL operation and the reliability of downstream surveying products.
To overcome this limitation, we present AWARE, a bio-inspired whole-body active yawing framework that exploits the UAV's own rotational agility to extend the effective sensor horizon and improve LIO's observability without additional mechanical actuation.
The core of AWARE is a differentiable Model Predictive Control (MPC) framework embedded in a Reinforcement Learning (RL) loop. It first identifies the viewing direction that maximizes information gain across the full yaw space, and a lightweight RL agent then adjusts the MPC cost weights online according to the current environmental context, enabling an adaptive balance between estimation accuracy and flight stability.
A Safe Flight Corridor mechanism further ensures operational safety within this HITL paradigm by decoupling the operator's navigational intent from autonomous yaw optimization to enable safe and efficient cooperative control. We validate AWARE through extensive experiments in diverse simulated and real-world environments. Our framework consistently outperforms passive scanning and static optimization baselines, achieving more accurate and robust LIO with real-time computational performance on resource-constrained UAV hardware. The project page can be found at: https://yizhezhang0418.github.io/aware.github.io/
\end{abstract}





\maketitle

\input{chapter/1Introduction}

\input{chapter/2RelatedWorks}


\input{chapter/4Method}

\input{chapter/5Simulation_Introduction}

\input{chapter/6Experiment}

\input{chapter/7Discussion}

\input{chapter/8Conclusion}


\printcredits

\section{Declaration of interests}
The authors declare that they have no known competing financial interests or personal relationships that could have appeared to influence the work reported in this paper.

\section{Acknowledgment}
This study was jointly supported by the National Natural Science Foundation Project (No. 42201477, No. 42130105) and Wuhan Natural Science Foundation Project (No. 2025041001010363).

\appendix

\bibliographystyle{cas-model2-names}

\FloatBarrier
\bibliography{reference}

\end{sloppypar}
\end{document}

%% file: chapter/1Introduction.tex
\section{Introduction}
\label{Introduction}

\begin{figure*}[!h]
\centering
\includegraphics[width=.98\textwidth]{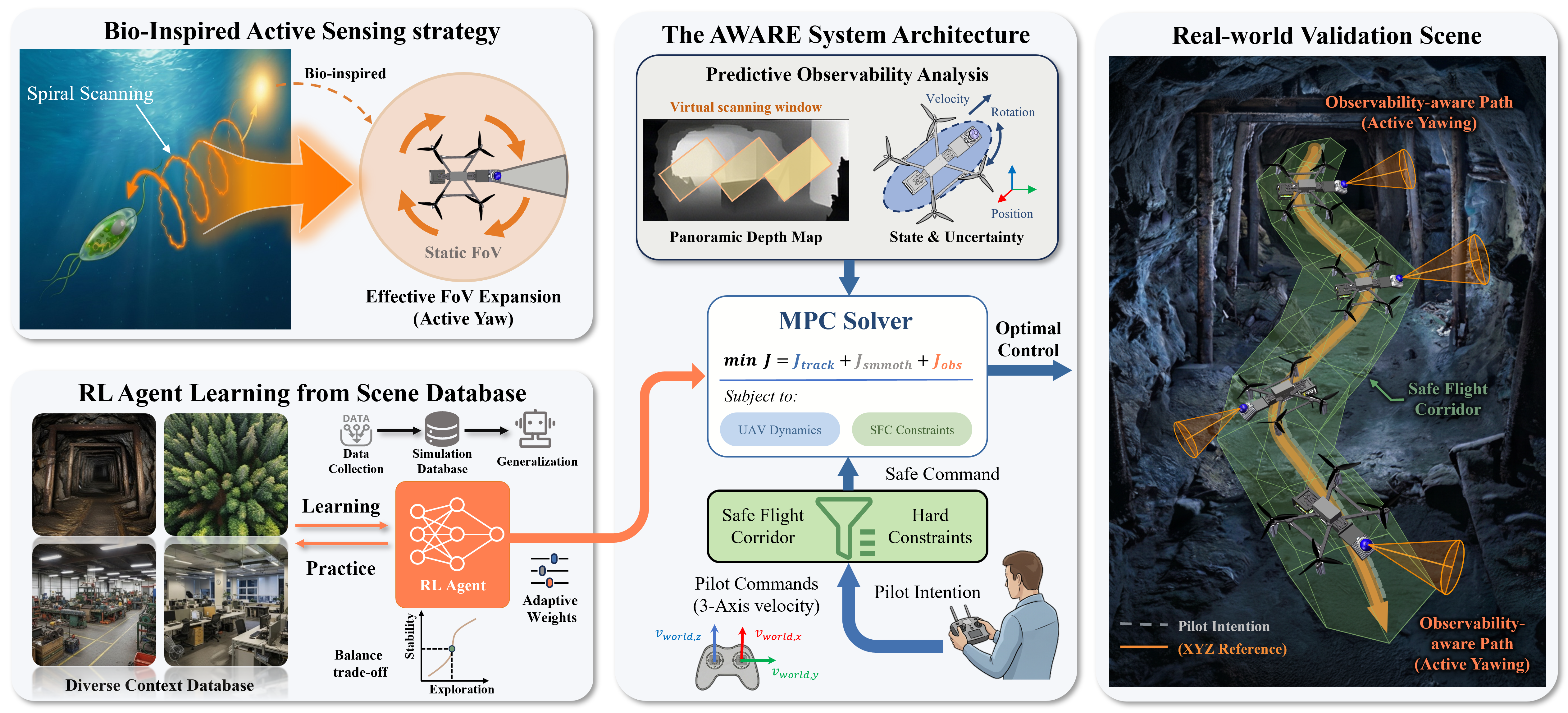}
\caption{Overview of the proposed AWARE framework. \textbf{Upper left:} Bio-inspired active sensing strategy that expands FoV through adaptive yaw control. \textbf{Lower left:} The RL agent learns adaptive cost weights from a diverse scene database, balancing observability and flight stability. \textbf{Center:} System architecture with predictive observability analysis and hybrid MPC, operating under human-in-the-loop operations and safe flight corridor constraints. \textbf{Right:} Real-world validation showing observability-aware yaw planning along the pilot's intended path.}
\label{fig:fig1}
\end{figure*}

Complex aerial surveying and robotics missions, such as underground mine inspection \citep{park2020applications, puniach2021application, ebadi2023present}, disaster response \citep{greenwood2019applications, khan2022emerging, liu2025artemis}, and dense forest surveying \citep{cui2014autonomous, li20193d, hao2026mapping}, represent some of the most demanding yet impactful application scenarios for UAVs. These environments are typically unstructured, safety-critical, and only partially observable, making fully autonomous operation difficult to rely on in practice. For such missions, the quality of the final surveying outcome depends fundamentally on whether the platform can maintain high-accuracy and geometrically reliable state estimation throughout the mission. Autonomous exploration strategies may fail to prioritize mission-relevant regions, while unexpected obstacles or environmental changes can quickly invalidate pre-planned trajectories \citep{zhang2025faem, zhang2025lidar}. As a result, human operators must remain involved in the loop to redirect the UAV toward regions of interest, supervise mission progress, and assess mapping completeness online \citep{zhang2020falco, ren2025survey}. In this sense, human-in-the-loop (HITL) control is not merely a practical add-on, but a fundamental requirement for real-world deployment in complex environments. However, such a HITL paradigm can only be effective when the UAV maintains reliable and accurate state estimation throughout operation, since degraded localization directly undermines both the safe execution of operator commands and the geometric fidelity of downstream surveying products \citep{cong2025adaptive}. LiDAR is particularly well suited for this role due to its accurate geometric perception and robustness under degraded or highly variable illumination conditions \citep{cadena2017past, zhao2021super, yin2024survey}. Yet UAV platforms operate under strict payload and power constraints, which preclude the use of heavy multi-sensor payloads \citep{chen2023self}. Consequently, they often rely on lightweight solid-state LiDARs with narrow fields of view (FoV) \citep{wang2021lightweight, xu2022fast}. This hardware limitation creates a fundamental challenge for precision aerial surveying: in geometrically degenerate or feature-sparse environments, restricted FoV significantly weakens the observability of LiDAR-inertial odometry (LIO), often leading to drift accumulation, unstable state estimation, and degraded geometric accuracy \citep{lin2020loam, zhang2024lio}. Such degradation not only compromises safe HITL operation, but also undermines the reliability of downstream mapping and scene understanding.


To achieve robust and high-accuracy LIO in geometrically degenerate environments, particularly given the restricted Field of View (FoV) of lightweight sensors, research has shifted towards active perception strategies \citep{ebadi2023present}. These approaches actively align sensor orientation with state-estimation objectives to maximize information gain and strengthen geometric constraints.
To address FoV limitations, existing solutions can be broadly categorized into hardware augmentation and algorithmic compensation. Initial attempts often utilized passive rotation strategies, where the sensor or platform rotates at a fixed rate to increase coverage \citep{alismail2015automatic, kaul2016continuous}. However, because these methods do not consider the environmental features, they lack the adaptability required to handle geometrically degenerate or feature-sparse environments and to sustain accurate odometry in practical surveying scenes \citep{chen2024design}. To introduce perception awareness, hardware-centric approaches, exemplified by UA-MPC \citep{li2025ua} and AEOS \citep{li2026aeos}, employ motorized mechanisms to actively sweep the sensor based on real-time uncertainty metrics. While effective, these motorized systems reduce flight endurance and increase mechanical vulnerability.
To avoid these hardware constraints, some concurrent research efforts have focused on optimizing the UAV's active motion via software-based control, typically utilizing Model Predictive Control (MPC) or end-to-end Reinforcement Learning (RL). However, significant challenges remain. Pure MPC frameworks usually rely on fixed cost weights, which are sensitive to parameter tuning and struggle to adapt across diverse environments \citep{jameson2012lockheed, zhu2025flare}. Conversely, end-to-end RL policies, while adaptable, often suffer from poor interpretability \citep{xu2025flying, zhang2026high}. Critically, neither category is designed with HITL operation in mind: they treat the UAV as a fully autonomous agent and provide no principled mechanism for safely integrating operator intent with autonomous yaw planning. This reveals a dual gap in current research: the absence of a purely software-driven approach that can (i) adaptively maintain accurate and robust LIO in changing environments without adding mechanical complexity, and (ii) seamlessly incorporate human operator commands into a safety-guaranteed, observability-aware control loop.

To bridge this gap, we propose \textbf{\emph{AWARE}}, a bio-inspired, control-and-observability-aware framework that employs adaptive whole-body active rotation for enhanced LiDAR-inertial odometry in complex surveying environments. Our inspiration comes from light-seeking microalgae in nature \citep{drescher2010fidelity}. These tiny organisms use a spiral swimming motion to continuously scan their surroundings, acting like a rotating antenna searching for light. By adjusting their movement patterns, they effectively steer toward optimal light sources. Similarly, our method enables the UAV to interactively adjust its yaw orientation and flight dynamics to sense the environment, thereby reducing localization uncertainty and improving estimation precision.
Unlike gimbal-based solutions that decouple sensing from flight, we introduce AWARE that uses the UAV's own agility to extend the effective sensor horizon without adding heavy mechanical components. However, aggressive whole-body rotation can destabilize flight and harm the odometry it aims to improve. The key insight of our method is to dynamically adapt the UAV's control model according to scene geometry and current estimation quality. This adaptation enhances LIO accuracy, robustness, and geometric reliability while maintaining stable and efficient flight operations.

To realize this adaptive capability and resolve the conflict between observability and stability, we design a hybrid architecture that combines the robustness and interpretability of differentiable Model Predictive Control (MPC) with the adaptability of Reinforcement Learning (RL). Our method operates on two levels. First, we build a panoramic observability predictor that evaluates geometric features from panoramic depth maps to identify the optimal viewing directions for state estimation. This guides the MPC to exploit informative regions. Second, a lightweight RL agent perceives the environmental context and dynamic state to adaptively model the control behavior that matches the current state. This hybrid RL-MPC design intelligently balances the trade-off between sensing gain and flight stability, enabling the UAV to execute adaptive flight control behaviors while maintaining accurate and robust odometry throughout the mission.
Furthermore, to address the lack of principled HITL support, we incorporate a Safe Flight Corridor (SFC) mechanism directly into the control architecture. Rather than treating human pilot commands as disturbances, AWARE decouples the operator's navigational intent from autonomous yaw optimization: the SFC constrains pilot commands to a collision-free corridor while the RL-MPC layer continues to optimize sensor observability in the background. This co-design enables safe and efficient cooperative control, where the human steers the mission-level trajectory and AWARE autonomously maximizes localization quality along that trajectory, thereby supporting reliable data acquisition in practical surveying operations. Together, active perception, adaptive stability control, and human-robot collaboration are unified into a single coherent framework.
Finally, to ensure robustness and transferability, we develop a high-fidelity, point cloud-based simulation platform that supports real-to-sim-to-real transfer across diverse complex scenes, including urban, underground, and forest environments. Our framework is trained and validated in simulation before deployment on a resource-constrained, whole-body yawing UAV platform. Our work makes the following contributions:

(1) We develop an Adaptive Whole-body Active Rotating Control framework that treats UAV yaw as a gimbal-like sensing action to improve LiDAR observability without additional mechanical actuation. A panoramic depth representation together with a Fisher Information Matrix-based metric enables the system to predict and actively steer toward viewing directions that reduce localization drift and improve odometry precision.

(2) We propose a Hybrid RL-MPC architecture that realizes observability-aware yaw control while balancing sensing gain and flight stability. A lightweight RL meta-controller dynamically adapts MPC cost weights to scene-dependent geometry, reducing manual tuning and enabling accurate and robust LIO across diverse environments.

(3) We integrate the controller into a human-in-the-loop setting, where operator commands are followed under autonomous yaw planning and Safe Flight Corridor constraints. This design preserves pilot intent and practical safety requirements while allowing observability-oriented yaw adaptation during precision surveying deployment.

(4) We develop a high-fidelity, point cloud-based simulation platform for training and evaluating observability-aware active yaw policies in closed loop. Built around realistic LiDAR rendering, LIO feedback, and diverse real-world maps, it supports systematic ablation, sim-to-real transfer, and real-world validation of LIO performance.

The remainder of this paper is organized as follows. Section~\ref{sec:related_works} reviews related work on active perception, LiDAR-based state estimation, and data-driven control strategies. Section~\ref{Methodology} details the proposed AWARE framework, including the observability-aware panoramic predictor and the hybrid RL-MPC architecture. Section~\ref{sec:simulation introduction} introduces the point cloud-based simulation platform and the control policy training pipeline. Section~\ref{sec:Experiments} reports experimental results in both simulated and real-world scenarios. Section~\ref{Discussion} discusses the findings and future directions. Finally, Section~\ref{sec:conclusion} concludes the paper.

%% file: chapter/2RelatedWorks.tex
\section{Related works}
\label{sec:related_works}


The evolution of active perception in aerial robotics represents a paradigm shift from mechanical complexity to algorithmic sophistication. While earlier methodologies sought to overcome sensor field-of-view (FoV) limitations by adding external actuation hardware, modern approaches are increasingly constrained by the rigorous Size, Weight, and Power (SWaP) limits of micro UAVs. Consequently, the field is moving towards a coupled flight-sensing paradigm, where the burden of expanding perceptual coverage is shifted from heavy mechanical gimbals to intelligent software that coordinates the vehicle's flight dynamics with its sensing objectives.

\subsection{From Limited Payload to Lightweight Whole-Body Perception Control}

To expand the restricted field-of-view (FoV) of statically mounted sensors, early systems heavily relied on mechanical actuation. Predominantly deployed on ground robots, these hardware-driven setups utilized motorized gimbals or rotating mounts to actively sweep the environment, thereby achieving denser 3D reconstructions and more robust SLAM performance in expansive areas \citep{zhang2014loam, alismail2015automatic, zhen2017robust}.
Recent studies have extended these designs to mobile platforms with relatively relaxed constraints, such as legged robots or systems utilizing commercial motorized scanners \citep{gong2023rss, li2025limo, livox2024scanner}. These setups allowed for high-frequency, omnidirectional scanning because the platforms could support the associated payload and power demands.

However, transferring these designs to UAVs introduces substantial challenges due to SWaP limitations. Consequently, most aerial platforms have relied on passive or mechanically constrained solutions, such as free-spinning mounts \citep{chen2023self} or MEMS-based optical deflection units \citep{mi11050456, juliano2022metasurface}, to improve coverage without heavy actuation. 
Critically, these systems lack the ability to actively direct sensing based on scene geometry or motion state. 
To bridge this gap, recent research has explored active sensing mechanisms to expand the limited field of view of aerial platforms. For example, FLARE \citep{zhu2025flare} dynamically adjusts its scanning direction using an actively rotated LiDAR. However, optimizing the highly nonlinear perception objective requires a Model Predictive Path Integral approach. The resulting reliance on extensive Monte Carlo sampling imposes a severe computational burden that can easily overwhelm lightweight UAVs.
Similarly, biologically inspired systems such as AEOS have developed active gaze control mechanisms akin to owls to dynamically reorient sensing toward salient regions \citep{li2026aeos}. While these motorized designs effectively enable environment-aware scanning, the addition of external rotating mechanisms inevitably introduces extra payload weight and mechanical complexity compared to traditional fixed-sensor configurations. For lightweight UAVs, the platform itself can instead be viewed as a natural gimbal: by exploiting the vehicle's own yaw motion, one can enlarge the effective sensing horizon without adding extra actuation hardware. This observation motivates coupled flight-sensing strategies that use whole-body yawing to improve viewpoint diversity and observability under strict onboard resource constraints.


\subsection{From Model-Based Constraints to Hybrid Learning Frameworks}


Traditional control methods, particularly Model Predictive Control (MPC), have long been the standard in aerial robotics due to their rigorous constraint handling and stability guarantees \citep{foehn2021time, romero2022model, romero2022time}. However, their reliance on manual tuning and fixed dynamic models limits performance in unknown or changing environments. To address these rigidities, Reinforcement Learning (RL) has emerged as a promising alternative, offering adaptive policies learned directly from interaction \citep{song2023reaching, xing2024contrastive}. Nevertheless, deploying pure RL for full flight control presents fundamental challenges: end-to-end policies often operate as opaque black boxes lacking the theoretical stability guarantees of traditional control \citep{xu2025navrl}, while processing high-dimensional sensory data demands heavy neural architectures that overwhelm lightweight onboard processors \citep{ling2023efficacy, zhang2025armor}, creating a difficult trade-off between adaptability and feasibility.

To bridge this gap, recent research has converged on hybrid architectures that integrate learned components into stable MPC frameworks. In ground robotics, Cao et al. \citep{cao2025learning} demonstrate that neural networks can predict optimal MPC weights for crowd navigation. Extending this to aerial robotics, Romero et al. \citep{romero2024actor, romero2025actor} successfully apply an actor-critic approach to tune MPC cost functions for agile flight on lightweight platforms. Similarly, for active sensing, AEOS \citep{li2026aeos} employs a learned policy to generate an implicit cost map that directly modulates the MPC objective to guide environmental exploitation and exploration, enhancing mapping quality while remaining efficient enough for aerial systems.

However, existing hybrid methods largely treat the environment as static geometric constraints, neglecting how specific scene features directly impact state estimation accuracy. Consequently, they lack the capability to actively regulate control maneuvers to mitigate localization drift, highlighting a critical need for frameworks that explicitly couple environmental modeling with robust flight control.

\subsection{Safe Active Sensing with Human Intent Preservation}

Ensuring safety during aggressive active sensing maneuvers is paramount for practical UAV operations. Two dominant paradigms exist for safe trajectory planning. The Euclidean Signed Distance Field (ESDF) approach \citep{oleynikova2017voxblox, han2019fiesta} provides continuous distance gradients that enable smooth, gradient-based trajectory optimization \citep{zhou2019robust}, with efficient ESDF-free variants further reducing computational cost \citep{zhou2020ego}. However, ESDF-based methods fundamentally enforce safety as a soft constraint through penalty terms, offering no strict collision-free guarantee during aggressive maneuvers. In contrast, Safe Flight Corridor (SFC) methods confine trajectories within certified obstacle-free convex polytopes \citep{liu2017planning}, with recent refinements enabling high-speed navigation via receding horizon strategies \citep{ren2022bubble, wang2022geometrically}. SFC provides formal collision-free certificates and naturally supports human-in-the-loop (HITL) intent preservation by anchoring safe corridors to the pilot's reference path. Despite these complementary strengths, both paradigms assume perfect state estimation. In practice, aggressive maneuvers in feature-poor environments induce localization drift that invalidates safety guarantees, making static geometric assumptions alone insufficient \citep{mellinger2011minimum, aucone2024synergistic, xu2022fast}.

Recent works have begun to address the coupling between perception and safety. Conservative strategies exclude unknown space from the traversable region \citep{liu2025slope} or employ reactive control against dynamic disturbances \citep{liu2023integrated, kong2021avoiding}. Perception-aware trajectory planning methods \citep{falanga2018pampc, bartolomei2021semantic} go further by incorporating visual feature quality directly into the planning objective. However, these approaches predominantly target camera-based features and assume fully autonomous operation.



Despite these advances, a critical integration gap persists. Existing safe planners and HITL mechanisms primarily focus on collision avoidance and intent preservation, while sensing quality remains largely decoupled from the control objective. In contrast, our work mainly targets LiDAR observability-aware yaw planning to reduce SLAM drift during aggressive flight, while retaining compatibility with HITL references and SFC-based safety constraints. In this sense, human intent preservation and corridor constraints serve as practical deployment conditions within which the controller operates, whereas the central technical objective is to use whole-body yawing to improve viewpoint quality for localization.


%% file: chapter/4Method.tex
\section{AWARE}
\label{Methodology}

\subsection{Problem Formulation and System Overview}
The proposed framework establishes a HITL active sensing architecture designed to maximize localization accuracy through adaptive yaw planning while preserving the operator's navigational authority. 
As illustrated in Fig. \ref{fig:fig1}, inspired by the light-seeking behavior of microalgae that continuously rotate to scan their surroundings, our method actively rotates the UAV platform to expand its FoV and focus on salient regions, thereby reducing localization uncertainty. The system operates on asynchronous LiDAR measurements and high-frequency IMU data, accepting real-time control commands from a human pilot. The backend performs robust state estimation via LiDAR-Inertial Odometry (LIO). 
Concurrently, a \textit{Predictive Observability Analysis Module} transforms real-time point clouds into a unified spherical representation to evaluate the prospective odometry gain across the entire yaw space. This module identifies the optimal viewing direction that minimizes pose uncertainty, effectively decoupling the burden of perception-aware maneuvering from the human operator.

To navigate the trade-off between sensing quality, motion stability, and human control intent in complex environments, we introduce a \textit{Hybrid Control Policy}. This policy leverages a Reinforcement Learning (RL) agent as a meta-controller to dynamically modulate the weighting costs of the Model Predictive Control (MPC) formulation based on extracted scene feature. By fusing explicit observability metrics with the implicit environmental context learned by the agent, the control node generates optimal actuation commands, ensuring high-fidelity SLAM performance within a Safe Flight Corridor (SFC) while tracking the operator's reference trajectory. Here, the SFC denotes a collision-free convex region (or a sequence of convex regions) extracted from the local map, and it is used to encode obstacle-avoidance requirements as linear position constraints in MPC.

\subsubsection{State and UAV Dynamics}
Leveraging the differential flatness property of quadrotor, we define the system state $\mathbf{x}_k \in \mathbb{R}^{10}$ and control input $\mathbf{u}_k \in \mathbb{R}^4$ at time step $k$ as:
\begin{equation}
    \mathbf{x}_k = [\mathbf{p}_k^T, \mathbf{v}_k^T, \boldsymbol{a}_k^T, \psi_k]^T, \quad \mathbf{u}_k = [\mathbf{j}_k^T, \dot{\psi}_k]^T, 
\end{equation}
where $\mathbf{p}_k, \mathbf{v}_k, \boldsymbol{a}_k \in \mathbb{R}^3$ denote the position, velocity, and acceleration in the world frame, while $\psi_k \in \mathbb{R}$ represents the yaw angle. The control commands comprise the translational jerk $\mathbf{j}_k \in \mathbb{R}^3$ and the yaw rate $\dot{\psi}_k \in \mathbb{R}$.
The system dynamics are modeled as a discrete-time linear time-invariant system:
\begin{equation}
    \mathbf{x}_{k+1} = \mathbf{A} \mathbf{x}_k + \mathbf{B} \mathbf{u}_k, 
\end{equation}
where the state transition matrix $\mathbf{A} \in \mathbb{R}^{10 \times 10}$ and the control matrix $\mathbf{B} \in \mathbb{R}^{10 \times 4}$ encapsulate a decoupled dynamic model. Specifically, it consists of a triple-integrator for the translational states and a first-order integrator for the yaw dynamics (i.e., $\psi_{k+1} = \psi_k + \dot{\psi}_k \Delta t$, with $\Delta t$ being the discretization interval). This formulation assumes constant control inputs within each step and deliberately neglects higher-order rotational dynamics, thereby ensuring the computational efficiency required for real-time onboard execution.

\begin{figure*}
\centering
\includegraphics[width=.98\linewidth]{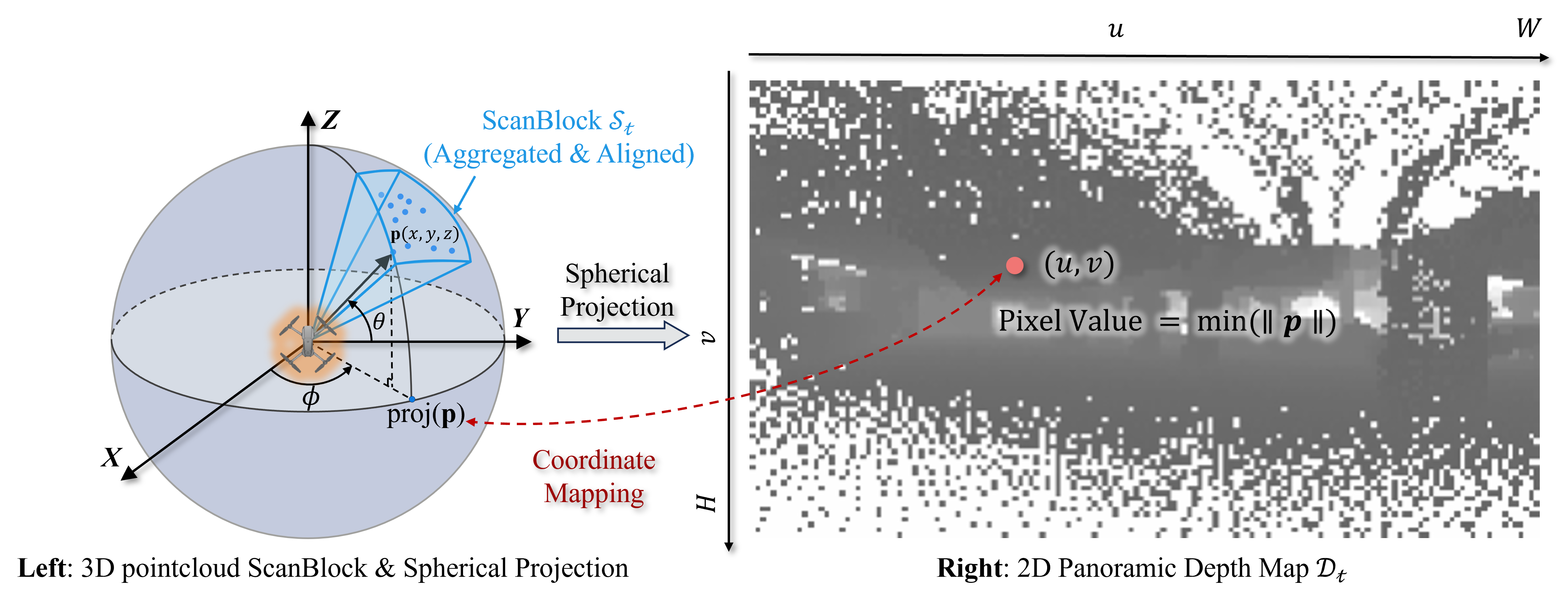}
\caption{Generation of the Unified Panoramic Representation. \textbf{Left:} The aggregated 3D pointcloud \textit{ScanBlock} $\mathcal{S}_t$ and the spherical projection model relative to the UAV body frame. \textbf{Right:} The resulting 2D Panoramic Depth Map $\mathcal{D}_t$. Each pixel $(u, v)$ corresponds to an angular direction and stores the minimum depth value, $\min(\| \mathbf{p} \|)$, to accurately represent the closest environmental obstacles.}
\label{fig:depthmap}
\end{figure*}

\subsubsection{Unified Panoramic Representation}
To efficiently integrate perception information into both the analytical observability analysis and the learning-based control module, we construct a unified representation of the local 3D scene, denoted as the Panoramic Depth Map $\mathcal{D}_t \in \mathbb{R}^{H \times W}$. As illustrated in Fig. \ref{fig:depthmap}, this formulation transforms the sparse 3D point cloud into a 2D grid structure, enabling efficient downstream processing while preserves the precise depiction of environmental characteristics.

High-quality mapping requires input data that is both dense and spatiotemporally consistent. To address the sparsity of single-frame LiDAR while adapting to the limited computational resources of the UAV platform, we construct the ScanBlock $\mathcal{S}_t$ by aggregating and downsampling a sliding window of recent observations. Historical scans are transformed into the current body frame to align with the present state:
\begin{equation}
    \mathcal{S}_t = \text{Downsample}\left( \bigcup_{k=0}^{N-1} \{ \mathbf{T}_{t, t-k} \cdot \mathbf{p} \mid \mathbf{p} \in P_{t-k} \} \right),
\end{equation}
where $\mathbf{T}_{t, t-k}$ aligns the past frame $P_{t-k}$ to the current body frame at time $t$. This process effectively densifies the geometric representation for robust perception while maintaining a manageable data size for real-time onboard processing.

The unified map $\mathcal{D}_t$ is then generated by projecting the aggregated points in $\mathcal{S}_t$ onto a discretized spherical coordinate system. For each point $\mathbf{p} = [x, y, z]^\top \in \mathcal{S}_t$, its pixel coordinates $(u, v)$ are derived via:
\begin{equation}
\begin{bmatrix} u \\ v \end{bmatrix} = 
\begin{bmatrix} 
2\pi + \frac{\arctan2(y, x)}{2\pi}  \cdot W \\ 
\pi - \frac{\arcsin(z / \|\mathbf{p}\|)}{\pi} \cdot H 
\end{bmatrix},
\label{eq:projection}
\end{equation}
\noindent where $H$ and $W$ represent the height and width of the panoramic map, respectively. The term $\|\mathbf{p}\| = \sqrt{x^2+y^2+z^2}$ denotes the Euclidean distance of the point from the sensor center. The functions $\arctan2(y, x)$ and $\arcsin(z / \|\mathbf{p}\|)$ compute the azimuth angle $\phi \in [-\pi, \pi]$ and the elevation angle $\theta \in [-\pi/2, \pi/2]$, which are then normalized to map onto the discrete pixel grid.

To capture precise occlusion patterns from this densified representation, each pixel $(u, v)$ encodes the shortest range observation within its angular bin:
\begin{equation}
    \mathcal{D}_t(u, v) = \min \{ \|\mathbf{p}\| \mid \mathbf{p} \in \mathcal{S}_t, \text{proj}(\mathbf{p}) = (u, v) \}.
\end{equation}

This representation serves as a shared interface: the high-resolution depth information supports the analytical uncertainty rendering in the MPC module, while a downsampled version acts as the state input for the RL network to guide trade-off between observability and UAV stability.

\subsubsection{Observability Analysis}

To quantify the localization quality, we analyze the observability of the LIO system based on the Fisher Information Matrix (FIM). The core of LIO relies on minimizing the point-to-plane distance residuals. For a point $\mathbf{p}_i$ in the body frame and its corresponding planar patch with normal $\mathbf{n}_i$ and centroid $\mathbf{q}_i$ in the global frame, the residual $\epsilon_i$ is defined as:
\begin{equation}
    \epsilon_i = \mathbf{n}_i^T (\mathbf{R}(\mathbf{x})\mathbf{p}_i + \mathbf{t}(\mathbf{x}) - \mathbf{q}_i),
\end{equation}
where $\mathbf{R}(\mathbf{x})$ is the rotation matrix and $\mathbf{t}(\mathbf{x})$ is the translation vector parameterized by the state $\mathbf{x}$. The FIM $\mathbf{\Phi}$ approximates the inverse of the estimation covariance matrix and is computed as:
\begin{equation}
    \mathbf{\Phi} = \sum_{i=1}^{N} \mathbf{J}_i^T \mathbf{J}_i,
\end{equation}
where $N$ is the number of point-to-plane correspondences, $\mathbf{J}_i = \partial \epsilon_i / \partial \mathbf{x}$ is the Jacobian of the residual with respect to the state.

\begin{figure}[]
    \centering
    \includegraphics[width=1.\linewidth]{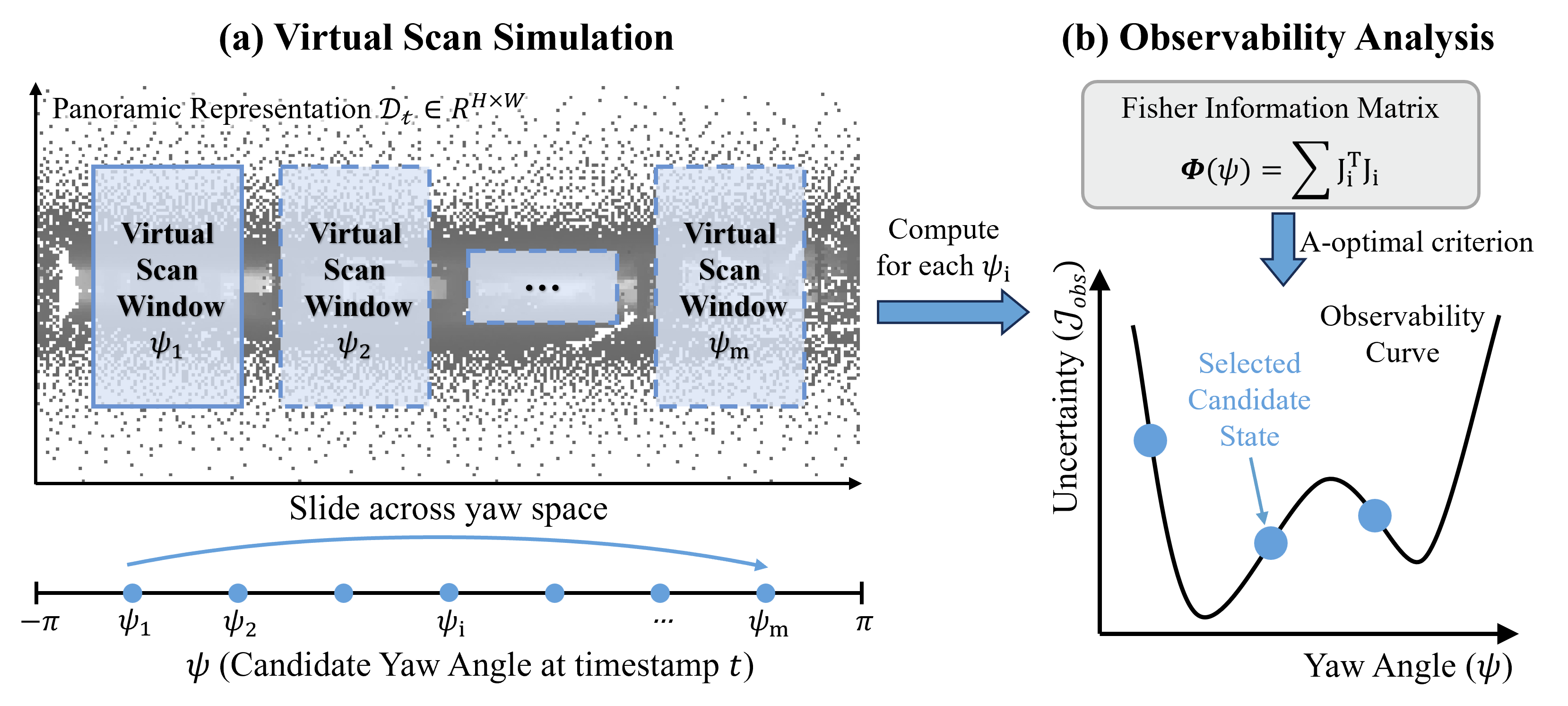}
    \caption{The Observability Analysis pipeline. We utilize the panoramic representation as input to simulate virtual scans for candidate yaw angles. By computing the Fisher Information Matrix (FIM) for each candidate, we generate an observability curve $\mathcal{J}_{obs}(\psi)$.}
    \label{fig:obs_pipeline}
\end{figure}

As illustrated in Fig. \ref{fig:obs_pipeline}, leveraging the panoramic representation $\mathcal{D}_t$, we evaluate the potential localization accuracy for a candidate yaw angle $\psi$ by virtually simulating the point cloud $P_t$ and computing the corresponding FIM $\mathbf{\Phi}(\psi)$. We adopt the A-optimality criterion as the observability metric $\mathcal{J}_{obs}$, which represents the average uncertainty of the estimation:
\begin{equation}
    \mathcal{J}_{obs}(\psi) = \text{Tr}(\mathbf{\Phi}(\psi)^{-1}),
\end{equation}
where $\text{Tr}(\cdot)$ denotes the matrix trace operator. Minimizing $\mathcal{J}_{obs}$ encourages the UAV to orient towards directions with rich geometric features that constrain the state estimation effectively.

While the observability analysis provides a scalar metric $\mathcal{J}_{obs}(\psi)$ for any given orientation, translating this perception objective into smooth, feasible motion requires a robust control framework. The following section details how this metric is seamlessly integrated into a HITL controller, transforming the passive evaluation of uncertainty into active, real-time trajectory optimization.

\subsection{Hybrid MPC Formulation with Human-in-the-Loop}
We formulate the control problem as a constrained Quadratic Programming (QP) optimization over a prediction horizon $N$. To ensure the system operates as a responsive assistant to the human pilot, the reference trajectory is directly coupled with the operator's real-time intent. 

Let $\mathcal{C}_{pilot} \in \mathbb{R}^4$ denote the raw commands from the remote controller (RC) joystick, mapping to longitudinal, lateral, vertical velocities, and yaw rate. In this HITL architecture, we implement a task decoupling strategy: the human operator retains primary authority over the translational motion ($x, y, z$) to ensure navigational intent, while the yaw dimension is autonomously optimized by the MPC to maximize observability.

This design is based on two key reasons. First, it enforces a clear division of labor: the human pilot focuses on the high-level navigational intent ("where to go"), while the system automatically manages the perceptual orientation ("how to look"). Second, LiDAR-based observability is predominantly sensitive to heading direction due to the sensor's FoV. Unlike translational motion, changes in yaw can drastically alter the visibility of texture-rich surfaces required for stable state estimation. Consequently, dedicating the yaw degree-of-freedom to active sensing allows the system to maximize SLAM fidelity without deviating from the operator's desired path.

The reference state $\mathbf{x}_{ref, k}$ at prediction step $k$ is generated by integrating the pilot's translational commands:
\begin{equation}
    \mathbf{x}_{ref, k} = \mathcal{T}_{int}(\mathbf{x}_{current}, \mathcal{C}_{pilot}, k \cdot \Delta t),
\end{equation}
where $\mathcal{T}_{int}$ represents the kinematic integration. This explicitly embeds the HITL nature of the system: the human acts as the high-level planner providing the guidance $\mathbf{x}_{ref}$, while the MPC locally refines the trajectory to satisfy constraints and actively plans the yaw angle for optimal perception.

The objective is to minimize a composite cost function subject to system dynamics, physical limits, and safety constraints:

\begin{align}
    \min_{\mathbf{u}_{0:N-1}} & \sum_{k=0}^{N-1} \Big( \lambda_{Q}\|\mathbf{x}_k - \mathbf{x}_{ref, k}\|^2 + \lambda_{R}\|\mathbf{u}_k\|^2 + \lambda_{R_{con}}\|\Delta \mathbf{u}_k\|^2 \nonumber \\
    & \quad + \lambda_{obs} \cdot \mathcal{J}_{obs}(\psi_k) \Big) + \lambda_{F}\|\mathbf{x}_N - \mathbf{x}_{ref}\|^2, \label{eq:mpc_cost} \\
    \text{s.t.} \quad & \mathbf{x}_{k+1} = \mathbf{A} \mathbf{x}_k + \mathbf{B} \mathbf{u}_k, \label{eq:mpc_dynamics} \\
    & \mathbf{x}_{min} \leq \mathbf{x}_k \leq \mathbf{x}_{max}, \quad \mathbf{u}_{min} \leq \mathbf{u}_k \leq \mathbf{u}_{max}, \label{eq:mpc_bounds} \\
    & \mathbf{A}_{sfc} \mathbf{p}_k \leq \mathbf{b}_{sfc}, \quad \forall k \in \{0, \dots, N-1\}, \label{eq:mpc_sfc}
\end{align}

\noindent where $\mathbf{x}_{ref,k}$ denotes the reference state at the $k$-th prediction step, dynamically generated from the human operator's real-time joystick inputs. $\Delta \mathbf{u}_k$ represents the control increment. The constraints include linearized dynamics (\ref{eq:mpc_dynamics}), actuation limits (\ref{eq:mpc_bounds}), and the Safe Flight Corridor (SFC) defined by $\mathbf{A}_{sfc}$ and $\mathbf{b}_{sfc}$ (\ref{eq:mpc_sfc}). The weighting coefficients in (\ref{eq:mpc_cost}) govern the trade-off between conflicting objectives: $\lambda_{Q}$ and $\lambda_{F}$ enforce trajectory tracking fidelity; $\lambda_{R}$ and $\lambda_{R_{con}}$ regularize control energy and smoothness; and $\lambda_{obs}$ incentivizes yaw behaviors that maximize state observability.

A critical challenge in integrating $\mathcal{J}_{obs}(\psi)$ into the differentiable MPC framework is that the observability metric lacks an analytical derivative with respect to the yaw angle $\psi$, making it incompatible with standard gradient-based QP solvers. To address this, we employ a quadratic approximation strategy.
Specifically, we sample the observability metric $\mathcal{J}_{obs}(\psi)$ at discrete yaw angles around the current orientation. We then fit a local quadratic function to these samples:
\begin{equation}
    \hat{\mathcal{J}}_{obs}(\psi) \approx \frac{1}{2} h_{obs} \psi^2 + g_{obs} \psi + c_{obs},
\end{equation}
where $h_{obs} $ and $g_{obs}$ represent the approximate Hessian and gradient of the observability landscape with respect to the yaw angle. These terms are directly incorporated into the MPC's objective function by augmenting the Hessian matrix $\mathbf{H}_{QP}$ and the gradient vector $\mathbf{g}_{QP}$ corresponding to the yaw state:
\begin{equation}
    \mathbf{H}_{QP}' = \mathbf{H}_{QP} + \lambda_{obs} h_{obs} \mathbf{I}_{\psi}, \quad \mathbf{g}_{QP}' = \mathbf{g}_{QP} + \lambda_{obs} g_{obs} \mathbf{e}_{\psi},
\end{equation}
where $\mathbf{I}_{\psi}$ is an indicator matrix that selects the yaw-related entries in the Hessian, and $\mathbf{e}_{\psi}$ is the corresponding unit vector for the gradient. This formulation allows the MPC to feel the curvature of the uncertainty landscape, effectively increasing the stiffness of the yaw controller in directions where observability is highly sensitive to orientation changes, while guiding the UAV towards the global minimum of uncertainty. The SFC constraints, denoted by $\mathbf{A}_{sfc}$ and $\mathbf{b}_{sfc}$, ensure that the optimized trajectory remains within the collision-free space defined by the map.

However, the performance of this MPC formulation relies heavily on the weighting coefficients. Fixed weights are ill-suited for dynamic and unstructured environments where the trade-off between exploration and exploitation changes rapidly. 
To address this, we propose a reinforcement learning framework to autonomously learn a state-dependent weighting policy, as detailed in the following section.

\subsection{RL-based Adaptive Parameter Tuning}

\begin{figure}[]
    \centering
    \includegraphics[width=.95\linewidth]{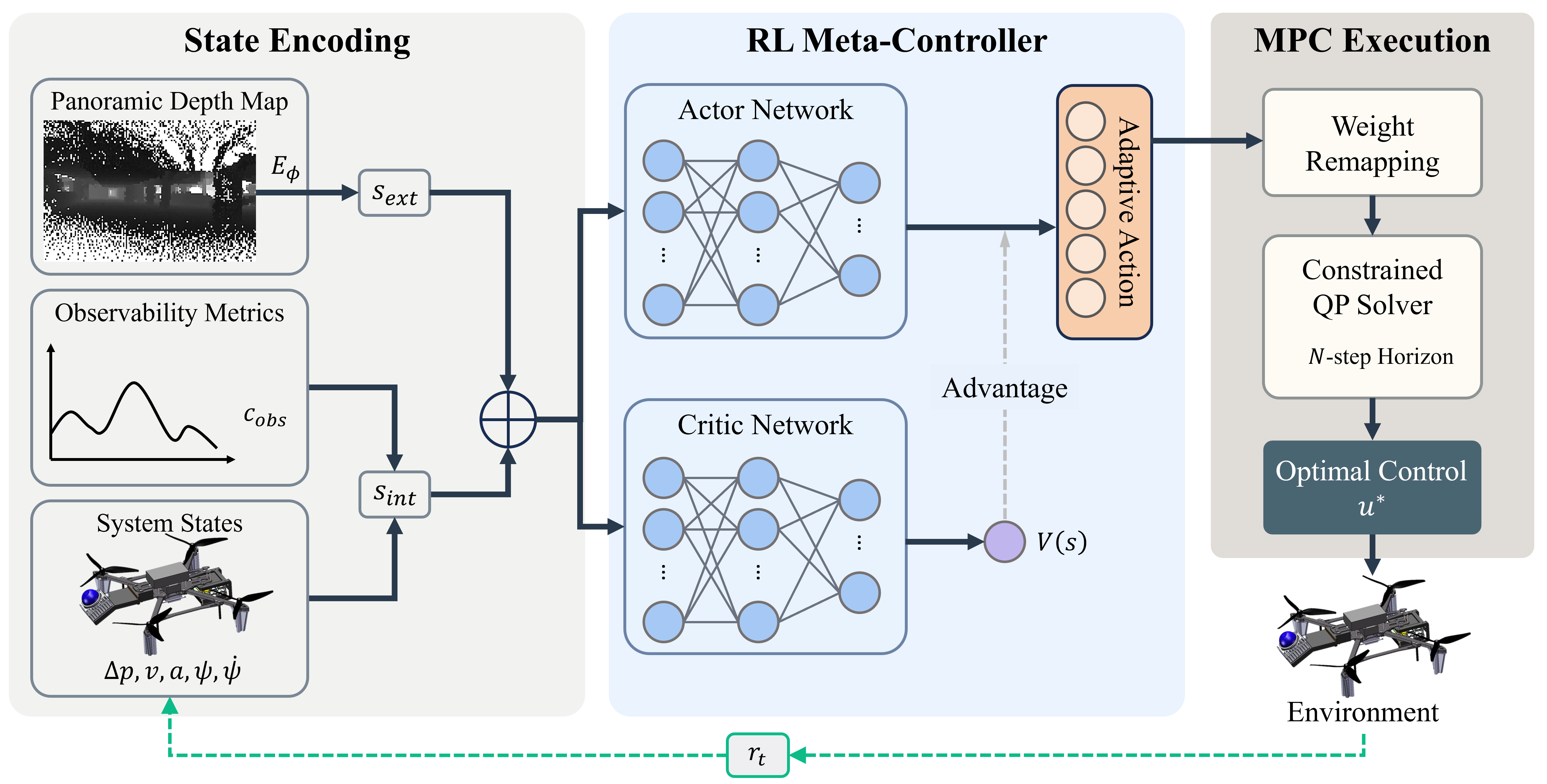}
    \caption{Overview of the Hybrid RL-MPC Control Architecture. The State Encoding module fuses UAV kinematics and observability metrics into $s_{int}$, and compresses the panoramic depth map into $s_{ext}$ via perception encoder $E_{\phi}$. The RL Meta-Controller uses an Actor-Critic network to output adaptive MPC cost weights. The MPC Execution module remaps these weights into a constrained QP solver to produce the optimal control command $u^*$. The reward $r_t$ closes the learning loop.}
    \label{fig:rl_architecture}
\end{figure}

To enable the UAV to dynamically adapt its control strategy across varying environments, we formulate the weight scheduling problem as a Partially Observable Markov Decision Process (POMDP). The RL agent acts as a meta-controller, observing the system state and environmental context to modulate the MPC cost weights in real-time. 

The architecture of this learning-based meta-controller is illustrated in Fig. \ref{fig:rl_architecture}. As depicted, the framework consists of three key components: (1) a hierarchical state encoding module that fuses proprioceptive kinematics with exteroceptive perception; (2) a lightweight Actor-Critic network optimized for real-time inference; and (3) an adaptive modulation interface that maps policy outputs to specific MPC weighting coefficients. This design effectively bridges the gap between high-level perception awareness and low-level control stability.

\subsubsection{State Space}
The state space $s_t$ is composed of an internal system state $s_{int}$ and an external perception state $s_{ext}$, defined as $s_t = [s_{int}, s_{ext}]$.

\textbf{1) Internal State:} The internal state vector $s_{int}$ encodes the UAV's kinematics and the local observability landscape:
\begin{equation}
    s_{int} = [\Delta\mathbf{p}_t, \mathbf{v}_t, \boldsymbol{a}_t, \psi_t, \dot{\psi}_t, \mathbf{c}_{obs}],
\end{equation}
where $\Delta\mathbf{p}_t$ represents the relative position of the UAV with respect to the goal position, $\mathbf{v}_t$ and $\boldsymbol{a}_t$ denote its velocity and acceleration, while $\psi_t$ and $\dot{\psi}_t$ indicate the yaw angle and yaw rate, respectively. The vector $\mathbf{c}_{obs}$ represents the sampled observability costs around the current yaw orientation.

\textbf{2) External State:} We utilize the panoramic depth map $\mathcal{D}_t$ to represent the local geometry. The map is flattened into a vector $\mathbf{v}_{\text{raw}} \in \mathbb{R}^{3200}$. 
To alleviate the computational burden on the RL agent and facilitate efficient feature abstraction, we process $\mathbf{v}_{\text{raw}}$ through a shallow MLP encoder $E_{\phi}$. This step maps the sparse, high-dimensional input to a compact and lightweight representation:
\begin{equation}
    \mathbf{s}_{\text{ext}} = E_{\phi}(\mathbf{v}_{\text{raw}}) \in \mathbb{R}^{128}.
\end{equation}
This 128-dimensional latent vector encapsulates the key environmental contexts and serves as the external observation for the policy network.

\rmk We employ a multi-resolution strategy for the spherical representation. A high-resolution depth map is utilized in the observability analysis module to ensure precise uncertainty gradient calculations. In contrast, for the RL agent's external state, we use a lower-resolution depth map to encode the environmental geometry. This design balances the need for precise observability analysis with the computational efficiency required for reinforcement learning, allowing the agent to effectively perceive the scene layout and identify navigable areas within a manageable state space.

\subsubsection{Network Architecture}
Given the deployment on resource-constrained onboard platforms, we design a lightweight network architecture to ensure real-time performance. The point cloud is projected into a spherical depth map to reduce dimensionality while preserving essential environmental features. 
The policy network employs an Actor-Critic architecture optimized with the Proximal Policy Optimization (PPO) algorithm. Specifically, the external state vector $s_{ext}$ is directly concatenated with the internal state vector $s_{int}$. This combined state vector is then fed into separate Multi-Layer Perceptron (MLP) networks for the Actor and Critic. 

We explicitly configure the hidden layers with sizes of 256, 256, and 128 units. This architecture strikes a balance between performance and efficiency: the initial deep layers provide sufficient capacity to capture complex non-linear dynamics, while the final layer acts as a bottleneck to condense features into a compact representation. This design enhances the robustness of action selection without incurring excessive computational overhead suitable for the resource-constrained UAV platform.

\subsubsection{Action Space}
The action space $\boldsymbol{ac}_t \in [-1, 1]^5$ is a continuous vector that modulates the diagonal weights of the MPC cost matrices. The normalized actions are mapped to the actual weight vector $\mathbf{W}_{M}$ via a linear scaling function:
\begin{equation}
    \mathbf{W}_{M} = \mathbf{W}_{\min} + \frac{\boldsymbol{ac}_t + \mathbf{1}}{2} \odot (\mathbf{W}_{\max} - \mathbf{W}_{\min}),
\end{equation}
where $\mathbf{W}_{\min}$ and $\mathbf{W}_{\max}$ are predefined lower and upper bounds for the weight vector, and $\odot$ denotes element-wise multiplication. The five tunable weights correspond to three groups of control objectives: $\lambda_p, \lambda_v, \lambda_a$ govern the penalty on position, velocity, and acceleration tracking errors; $\lambda_{\dot{\psi}}$ regularizes the yaw rate for motion smoothness; and $\lambda_{obs}$ scales the observability cost in the MPC objective. Through this parameterization, the agent can continuously adjust the relative priority among trajectory tracking accuracy, rotational smoothness, and active sensing behavior in response to the evolving scene context.

\subsubsection{Reward Design}
The reward function $r_t$ combines three objectives that jointly shape the agent's policy: SLAM accuracy, exploration completeness, and control smoothness. Their weighted sum is given by:
\begin{equation}
    r_t = \alpha_1 r_{acc} + \alpha_2 r_{exp} + \alpha_3 r_{smooth}.
\end{equation}
We describe each component below.

The SLAM accuracy reward $r_{acc}$ incentivizes precise local state estimation. To decouple the evaluation from accumulated global drift, we apply Umeyama alignment to register the estimated trajectory with the ground truth within a local sliding window \citep{messikommer2024reinforcement}. The reward adopts an exponential kernel:
\begin{equation}
    r_{acc} = \exp\left(-300 \cdot \text{RTE}(\mathbf{p}_{est}, \mathbf{p}_{gt})\right),
\end{equation}
where $\text{RTE}(\cdot)$ denotes the Relative Trajectory Error after alignment. This form yields high sensitivity in the low-error region, encouraging highly precise local state estimation, while decaying rapidly as the error grows.

The exploration reward $r_{exp}$ encourages the discovery of unmapped areas and is computed as the ratio of newly observed voxels $N_{new}$ to the total observed voxels $N_{total}$. A square root is applied to moderate the magnitude and simplify weight tuning:
\begin{equation}
    r_{exp} = \sqrt{\frac{N_{new}}{N_{total}}}.
\end{equation}

The smoothness reward $r_{smooth}$ penalizes erratic yaw motions. Smooth rotation helps maintain sufficient FoV overlap for scan-to-map registration, which is critical for the convergence stability of the Kalman filter. It is defined via the variance of the yaw rate $\dot{\psi}$ over a sliding window of length $k$:
\begin{equation}
    r_{smooth} = \exp\left( -\text{Var}(\dot{\psi}_{t-k:t}) \right).
\end{equation}




%% file: chapter/5Simulation_Introduction.tex
\section{Point Cloud-Based Simulation and Control Policy Training}
\label{sec:simulation introduction}

\subsection{High-Fidelity Simulation Environment}
\begin{figure}[]
    \centering
    \includegraphics[width=.98\linewidth]{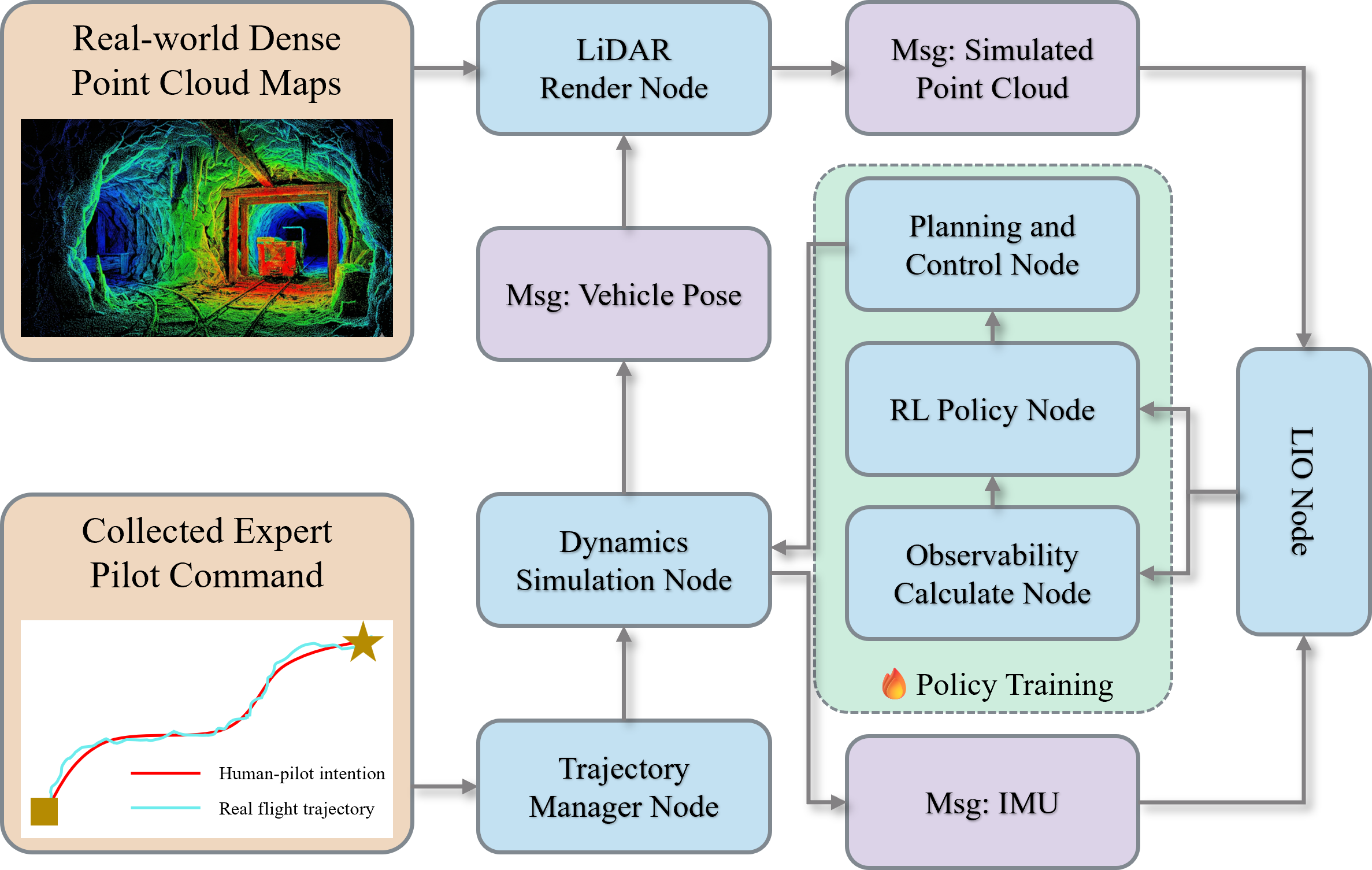}
    \caption{Simulation pipeline for the proposed active sensing framework. The environment integrates real-world point clouds and expert trajectories to drive UAV and LiDAR simulations. Within this closed loop, observability metrics and LIO feedback continuously train the RL policy to output control commands that maximize state estimation robustness.}
    \label{fig:simulator}
\end{figure}

To rigorously train and validate the proposed active sensing framework, we established a high-fidelity, closed-loop simulation environment, as shown in Fig. ~\ref{fig:simulator}. Adopting a Real-to-Sim-to-Real methodology, our platform synthesizes sensor data directly from dense point cloud maps collected in diverse real-world scenarios rather than relying on synthetic geometric primitives. This data-driven design provides two key training advantages: (i) learning realistic scanning patterns under real sensor noise and occlusion statistics, and (ii) fast policy warm-up through imitation of expert pilot behavior in the same geometric contexts.

The simulation architecture mirrors the information flow of a real-world UAV system, designed to facilitate seamless Sim2Real transfer for whole-body control tasks. The workflow initiates by feeding the collected sequence of expert pilot commands via the \textit{Trajectory Manager Node} into the \textit{Dynamics Simulation Node}. Crucially, this node incorporates realistic aerodynamic damping and control delays to emulate the physical response of the quadrotor. Beyond trajectory replay, this expert-command stream serves as a behavioral prior for policy learning, allowing the agent to inherit human-like scan-control rhythms during the early stage and avoid inefficient random exploration.

In the perception loop, the \textit{LiDAR Render Node} utilizes point cloud maps collected from physical environments as the base map to perform ray-casting based on the UAV's instantaneous 6-DoF pose. The resulting simulated point clouds, along with IMU data, are fed into the \textit{LIO Node} for state estimation. Subsequently, the estimated state and metrics from the \textit{Observability Calculate Node} serve as the input state space for the \textit{RL Policy Node}. The policy then outputs high-level actions to the \textit{Planning and Control Node}, which generates precise control commands for the dynamics simulator, thereby establishing a closed feedback loop.

Overall, the simulator is not only a realism-oriented rendering engine but also a data-driven training accelerator. By coupling real-world geometry with expert pilot behavior, it improves both the final Sim2Real robustness and the early-stage optimization efficiency of the policy.

\subsection{Training Datasets and Scenarios}

\begin{figure*}[]
    \centering
    \includegraphics[width=.95\linewidth]{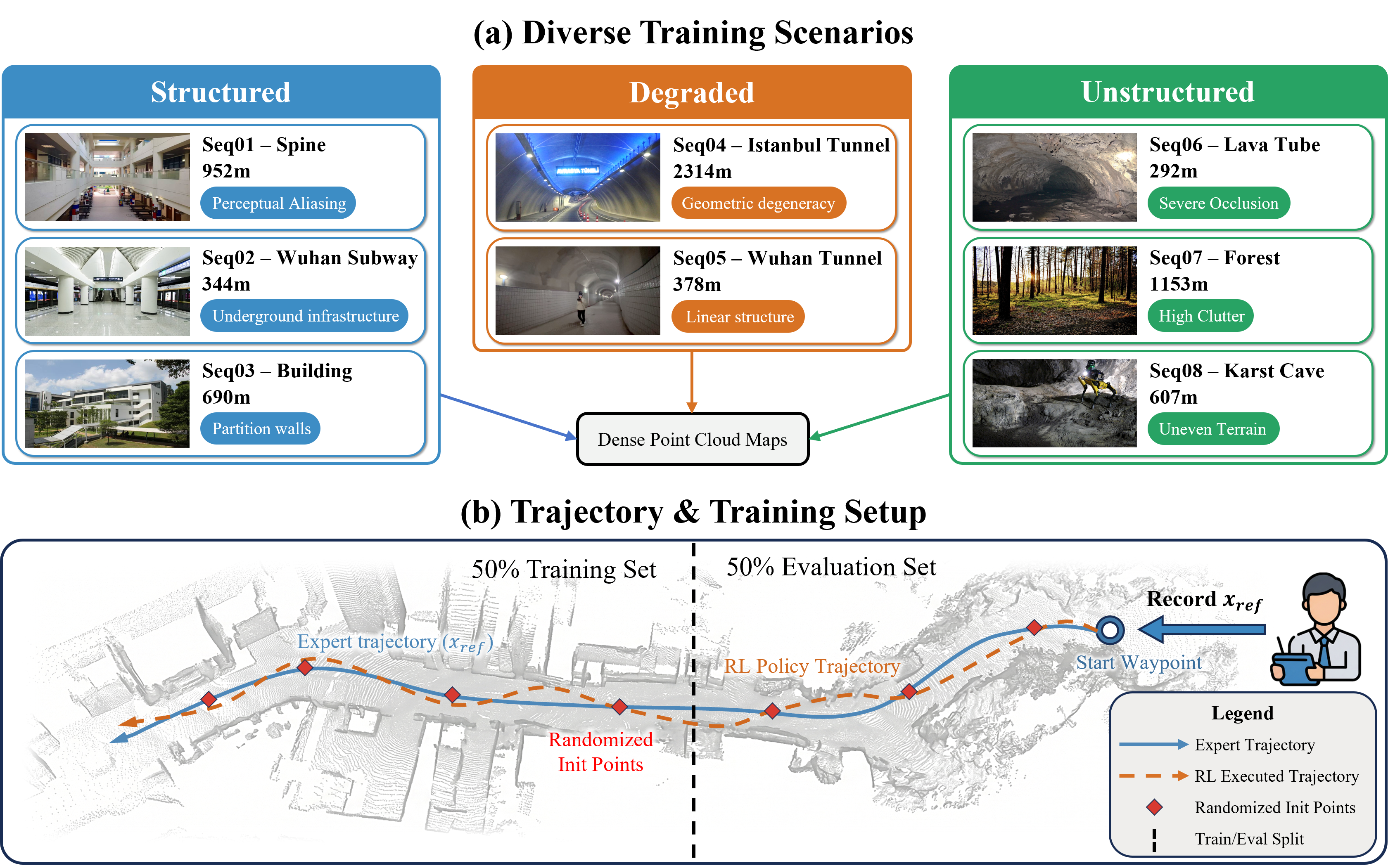}
    \caption{Overview of the simulation dataset and training configuration. (a) Diverse training scenarios spanning structured, degraded, and unstructured environments, from which dense point cloud maps are collected. (b) Trajectory and training setup: expert-demonstrated reference trajectories ($\mathbf{x}_{ref}$) serve as navigation baselines, with randomized initialization points along the path; the first 50\% of each trajectory is used for training and the remainder for evaluation.}
    \label{fig:scenarios}
\end{figure*}

Training a robust active sensing policy requires exposure to diverse and challenging environments. To this end, we constructed a dataset consisting of two components: dense point cloud maps and expert-demonstrated reference trajectories, as illustrated in Fig.~\ref{fig:scenarios}.

The point cloud maps are sourced from a combination of established public benchmarks \citep{li2023whu, petravcek2021large, autowarefoundation, dong2020registration} and several self-collected datasets. Together, these encompass a broad range of real-world operational environments, covering confined subterranean corridors, unstructured natural terrains, and large-scale man-made infrastructures. We detail the specific characteristics of each scenario in Table~\ref{tab:scenarios}. These environments impose distinct perceptual challenges on LIO systems, including severe geometric degeneracy, high-frequency clutter, and limited structural features that typically induce state estimation drift. By training across such diverse conditions, we prevent the policy from overfitting to any single scene type and encourage the learning of environment-agnostic sensing strategies.

We also provide ground truth trajectories recorded by human experts as coarse navigation baselines ($\mathbf{x}_{ref}$). These trajectories play a dual role. First, they reduce the complexity of long-horizon navigation, allowing the RL agent to focus on local trajectory refinement. Second, they provide action priors (e.g., viewpoint transitions and scan rhythm) that enable fast warm-up through behavior imitation. The agent then learns to actively adjust its 6-DoF pose to maximize SLAM observability while managing collision avoidance and trajectory tracking. During training, the initialization points are randomized along these expert trajectories, forcing the agent to generalize across heterogeneous geometric conditions.

\begin{table*}[htbp]
    \centering
    \renewcommand{\arraystretch}{1.1} 
    \caption{Detailed specification of the diverse simulation scenarios selected for active scanning training. The dataset is categorized into Structured, Degraded, and Unstructured environments, each presenting unique geometric challenges ranging from featureless tunnel to high-frequency clutter, ensuring comprehensive policy generalization.}
    \label{tab:scenarios}
    \begin{tabular*}{\textwidth}{@{\extracolsep{\fill}} lcllc @{}}
        \toprule
        \textbf{Sequence} & \textbf{Scenario Name} & \textbf{Category} & \textbf{Geometric Challenges} & \textbf{Length (m)} \\
        \midrule
        Seq01 & Spine              & Structured   & Linear corridor, repetitive vertical columns & 952 m \\
        Seq02 & Wuhan Subway       & Structured   & Large-scale indoor, repetitive patterns      & 344 m \\
        Seq03 & Building           & Structured   & Multi-level complex, partition walls         & 690 m \\
        \midrule
        Seq04 & Istanbul Tunnel    & Degraded     & Featureless Tunnel, geometric degeneracy   & 2314 m \\
        Seq05 & Wuhan Tunnel       & Degraded     & Unexposed space, linear structure             & 378 m \\
        \midrule
        Seq06 & Lava Tube          & Unstructured & Irregular curvature, severe occlusion        & 292 m \\
        Seq07 & Forest             & Unstructured & High clutter, unstructured vegetation        & 1153 m \\
        Seq08 & Nebra Karst Cave   & Unstructured & Complex topology, uneven terrain             & 607 m \\
        \bottomrule
    \end{tabular*}
\end{table*}

\subsection{Training Implementation}

Following the POMDP formulation in Sec.~\ref{Methodology}, we train the policy using Proximal Policy Optimization (PPO) for a total of $3 \times 10^6$ timesteps. To ensure robust generalization, we employ a diverse environment sampling strategy where the trajectory manager dynamically cycles through a spectrum of scenarios, including open structured spaces and geometrically degraded unexposed scenes. To ensure rigorous validation on unseen geometries, only the first 50\% of each expert trajectory is designated as the training set, explicitly reserving the remaining segments for evaluation.
In each episode, we randomly sample a 120-second segment. This duration is long enough for drift to accumulate, forcing the agent to learn how to correct it and maintain stability.
Additionally, we apply online reward normalization using Welford's algorithm to standardize the reward scale, which is crucial for stabilizing the value function estimation.

The optimization is performed using the Adam optimizer with a linearly decaying learning rate from $3 \times 10^{-4}$ to $1 \times 10^{-4}$. Future rewards are discounted with a factor of $\gamma = 0.99$, and Generalized Advantage Estimation (GAE) is applied with $\lambda = 0.95$. To encourage monotonic improvement and training stability, we enforce a clipping range of $\epsilon = 0.2$ and strictly limit the KL divergence between consecutive policy updates to 0.01. The policy is updated every 1024 rollout steps with a batch size of 512 across 8 optimization epochs. Furthermore, the loss function incorporates a value function coefficient of 0.6 and an entropy coefficient of $1 \times 10^{-4}$ to encourage sufficient exploration during the early stages of training.

%% file: chapter/6Experiment.tex
\section{Experiments}
\label{sec:Experiments}

\subsection{Evaluation Metrics and Baselines}
\label{sec:experiments_setup}

To rigorously assess the localization fidelity and robustness of the proposed framework, we adopt the Absolute Pose Error (APE) as the principal performance indicator. The APE quantifies the Euclidean deviation between the estimated trajectory and the ground-truth poses. A lower APE signifies that the active scanning strategy has successfully identified and tracked geometrically salient features, thereby minimizing drift in the LiDAR-Inertial Odometry (LIO) state estimation.
We benchmark the proposed adaptive framework against two categories of baselines: passive fixed-rate scanning and static optimization-based MPC.

The first category, passive fixed-rate scanning, represents conventional strategies that operate without environmental awareness. We evaluate two variants within this group. In the slow variant, the LiDAR rotates at a constant desired angular velocity of $\omega = 1.0$~rad/s, which yields high point density but often leads to insufficient spatial coverage during dynamic UAV flights. In the fast variant, the desired angular velocity is set to $\omega = 8.0$~rad/s to maximize Field of View (FoV) coverage, though this comes at the cost of sparser point distribution and increased motion distortion.

The second category, static optimization-based MPC, is designed to ablate the contribution of the RL-based adaptive tuning mechanism. Here, we compare our method against standard MPC formulations in which the observability weight $\lambda_{obs}$ is held constant throughout the trajectory. We test five values: $\lambda_{obs} \in \{100, 500, 1000, 1500, 2000\}$. Since these configurations enforce a fixed sensing priority regardless of the scene geometry, they cannot adapt to varying environmental conditions and thus serve as a natural reference for evaluating the benefit of dynamic weight adjustment.

To ensure a fair comparison, all experimental trials share the same simulation environment, LiDAR sensor parameters, UAV dynamic constraints, and initial state configurations. The reported results represent the average performance over the entire trajectory length for each sequence.

\subsection{Evaluation in Simulation}
\label{sec:sim_eval}

\begin{figure*}[!htbp]
\centering
\includegraphics[width=0.98\textwidth]{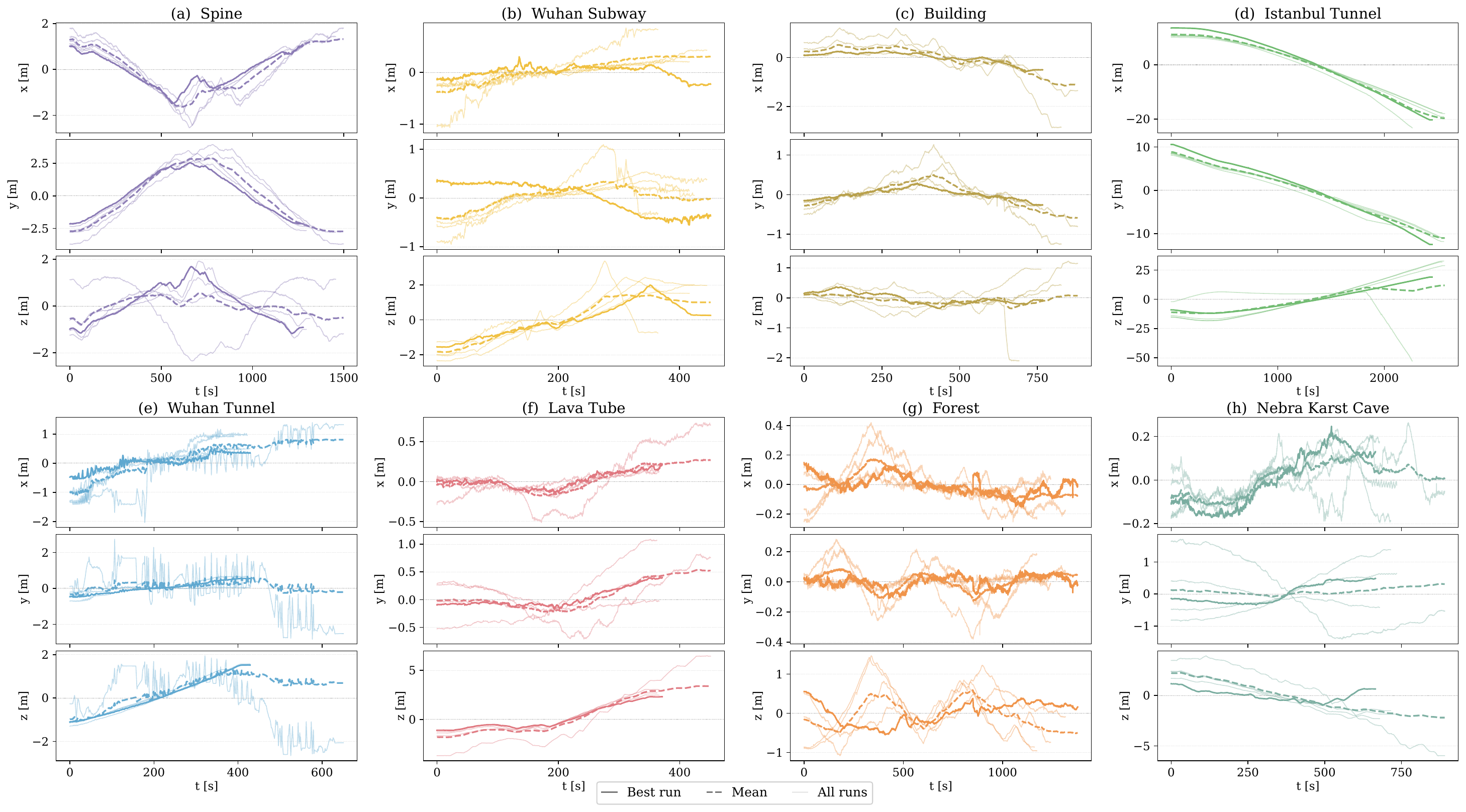}
\caption{Trajectory errors on the eight sequences of the AWARE dataset evaluated against the ground truth across five independent runs.}
\label{fig:all_scenes_xyz_err}
\end{figure*}

\begin{figure*}[]
\centering
\includegraphics[width=0.98\textwidth]{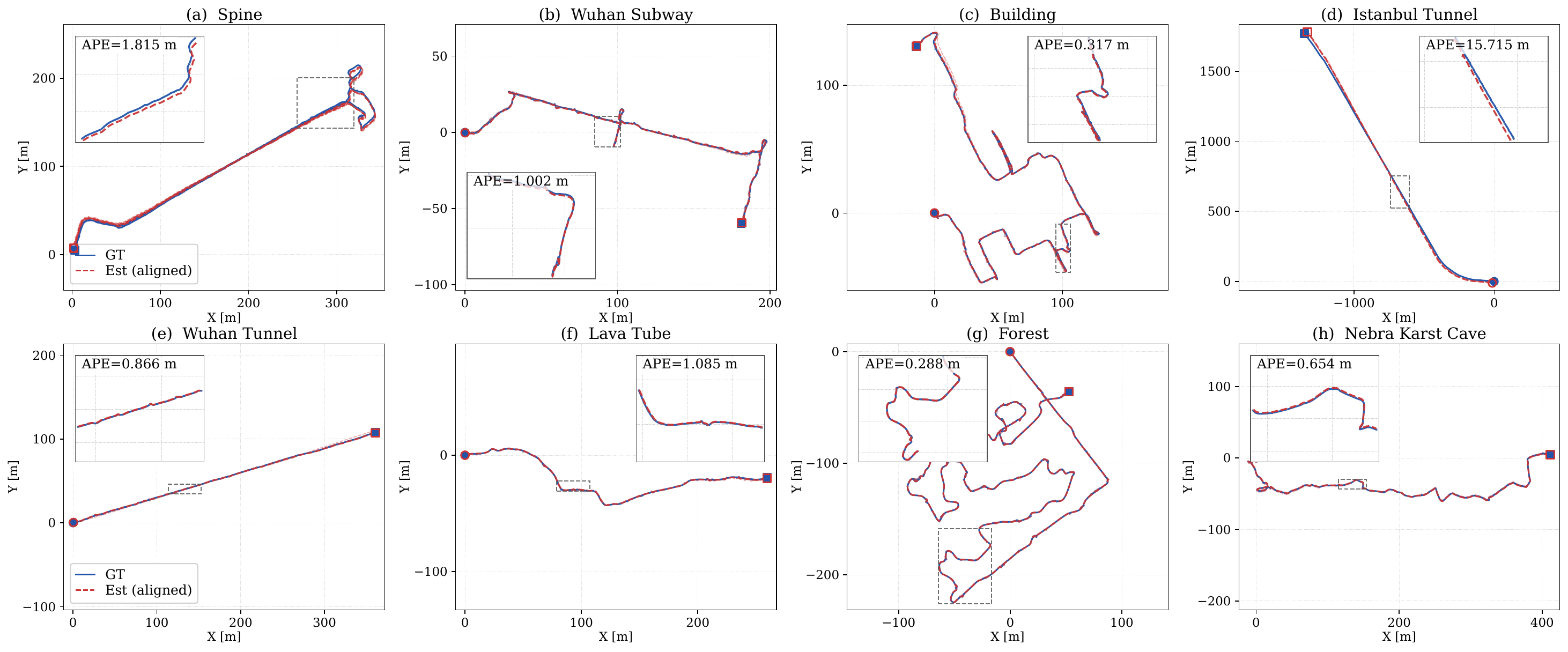}
\caption{Estimated trajectories on our simulation datasets aligned with the ground truth. The plots display the single run with the minimum Absolute Pose Error (APE) for each sequence. }
\label{fig:all_scenes_traj}
\end{figure*}

\begin{figure}[]
\centering
\includegraphics[width=0.98\linewidth]{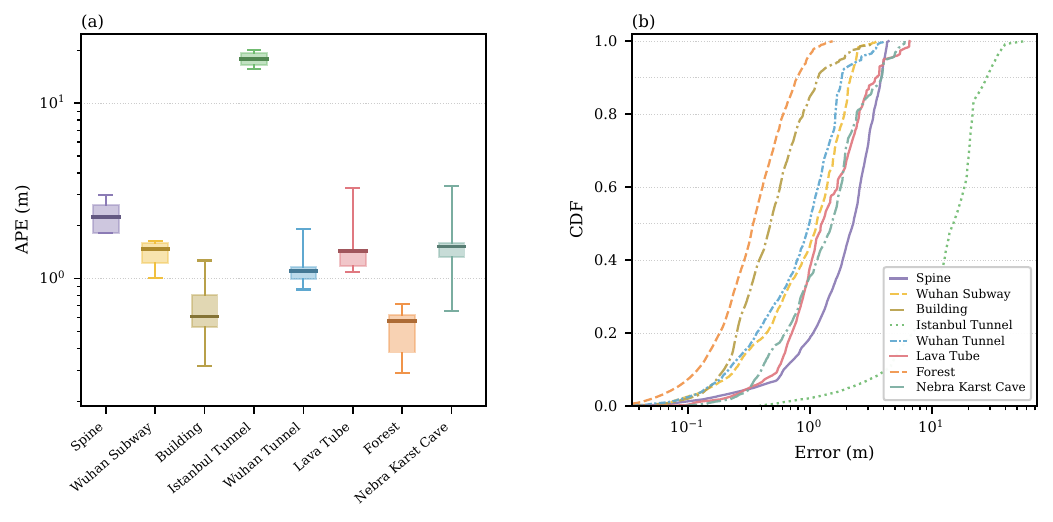}
\caption{Quantitative localization results of the proposed AWARE framework across eight diverse testing scenarios. (a) Box plots illustrating the distribution of the Absolute Pose Error (APE) in meters, presented on a logarithmic scale. (b) Cumulative Distribution Function (CDF) of per-frame trajectory error.}
\label{fig:all_scenes_boxplot}
\end{figure}


We first evaluated AWARE across the multiple diverse scenes mentioned previously. 
To validate the stability and repeatability of our adaptive formulation, we conduct five independent trials for each sequence. Fig.~\ref{fig:all_scenes_xyz_err} presents the per-axis trajectory errors of all five runs against the ground truth across the eight sequences, providing a detailed view of the estimation consistency. Fig.~\ref{fig:all_scenes_traj} visualizes the best-performing run (i.e., the one achieving the lowest APE) for each sequence, with the estimated trajectory overlaid on the ground truth. In geometrically rich environments such as \textit{Forest} and \textit{Building}, AWARE closely tracks the reference trajectory with sub-meter APE, confirming that the RL-guided scanning strategy effectively exploits abundant geometric features. Even in structurally degenerate scenes like \textit{Istanbul Tunnel} and \textit{Wuhan Tunnel}, the estimated trajectories remain qualitatively aligned with the ground truth, demonstrating the robustness of the active perception mechanism.

To further characterize the statistical distribution of localization errors, Fig.~\ref{fig:all_scenes_boxplot}(a) depicts box plots of the APE on a logarithmic scale. AWARE achieves sub-meter median APE in geometrically rich scenes and maintains bounded error in degenerate environments. Fig.~\ref{fig:all_scenes_boxplot}(b) plots the Cumulative Distribution Function (CDF) of per-frame trajectory errors. The steeper, left-shifted curve of AWARE indicates smaller localization errors for most frames, further attesting to the framework's overall reliability.

The aggregated results are reported in Table~\ref{tab:ape_results}. Overall, AWARE yields an average APE of 3.41~m, representing a 27.8\% reduction over the best static MPC baseline ($\lambda_{obs}=500$, 4.72~m) and an 84.9\% improvement over passive fixed-rate scanning.

Fixed-rate strategies exhibit a clear trade-off between point density and spatial coverage. Slow scanning provides dense measurements but limited observability, while fast scanning maximizes the field of view at the cost of sparse point distribution and motion distortion. Both variants fail in geometrically degenerate settings such as the \textit{Istanbul Tunnel}, where the lack of active perception leads to catastrophic drift with errors exceeding 100~m.

The static MPC baselines provide more adaptive behavior by incorporating observability objectives into the optimization. However, their performance is highly sensitive to the choice of the weight $\lambda_{obs}$. As shown in the table, the optimal weight varies across scenes: $\lambda_{obs}=1000$ works best for \textit{Spine}, while $\lambda_{obs}=500$ is preferred for \textit{Wuhan Subway} and $\lambda_{obs}=1500$ for \textit{Wuhan Tunnel}. No single configuration yields consistently good results across all environments. In contrast, AWARE leverages the RL agent to dynamically adjust the optimization weight based on real-time scene context. This enables the framework to automatically identify the optimal trade-off between trajectory tracking and active perception for each environment. As a result, AWARE consistently outperforms the best static configuration without requiring manual tuning, demonstrating strong generalization across structurally diverse scenes.

\begin{table*}[htbp]
  \centering
  \caption{APE (in meters) of different scanning control methods across simulated scenes. Lower is better. Bold indicates best performance. Underline indicates second-best.}
  \label{tab:ape_results}
  \resizebox{\textwidth}{!}{
  \begin{tabular}{llcccccccc}
    \toprule
    \textbf{Sequence} & \textbf{Scene} & \textbf{Fixed-rate (Fast)} & \textbf{Fixed-rate (Slow)} & $\mathbf{\lambda_{obs} = 100}$ & $\mathbf{\lambda_{obs} = 500}$ & $\mathbf{\lambda_{obs} = 1000}$ & $\mathbf{\lambda_{obs} = 1500}$ & $\mathbf{\lambda_{obs} = 2000}$ & \textbf{AWARE} \\
    \midrule
    Simu-Seq01 & Spine & 5.24 & 6.96 & 4.16 & 3.39 & \underline{3.04} & 3.49 & 3.35 & \textbf{2.29} \\
    Simu-Seq02 & Wuhan Subway & 14.80 & 8.41 & 5.26 & \underline{3.39} & 6.55 & 4.23 & 6.27 & \textbf{1.38} \\
    Simu-Seq03 & Building & 2.38 & 2.03 & 1.24 & 1.41 & 1.34 & \underline{1.18} & 1.23 & \textbf{0.71} \\
    Simu-Seq04 & Istanbul Tunnel & 139.31 & 132.89 & 21.70 & \underline{19.36} & 20.75 & 19.61 & 21.70 & \textbf{17.86} \\
    Simu-Seq05 & Wuhan Tunnel & 3.01 & 2.74 & 2.02 & 1.94 & 1.63 & \underline{1.42} & 1.62 & \textbf{1.20} \\
    Simu-Seq06 & Lava Tube & 3.92 & 3.87 & 3.05 & 3.20 & 3.28 & \underline{3.00} & 3.12 & \textbf{1.68} \\
    Simu-Seq07 & Forest & 1.29 & 1.32 & 0.89 & \underline{0.61} & 0.71 & 0.83 & 0.81 & \textbf{0.50} \\
    Simu-Seq08 & Nebra Karst Cave & 10.25 & 7.50 & 5.36 & 4.46 & \underline{3.31} & 4.89 & 4.25 & \textbf{1.69} \\
    \midrule
    \textbf{Average} & -- & 22.53 & 20.72 & 5.46 & \underline{4.72} & 5.08 & 4.83 & 5.29 & \textbf{3.41} \\
    \bottomrule
  \end{tabular}
  }
\end{table*}

\subsection{Real-World Deployment Validation}
\begin{figure}[]
\centering
\includegraphics[width=0.98\linewidth]{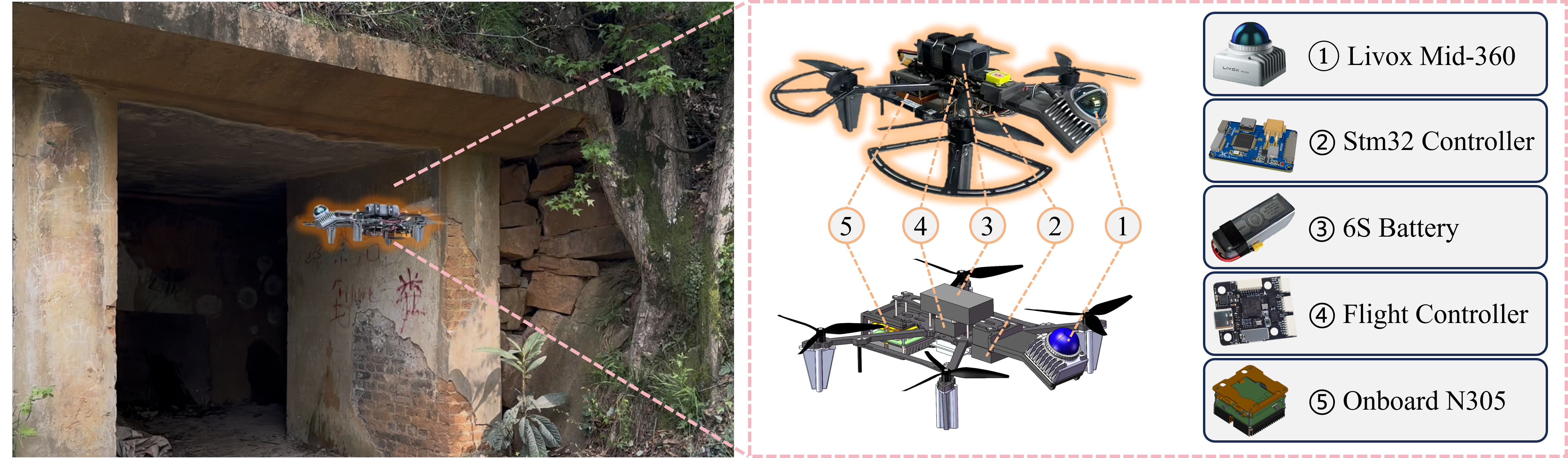}
\caption{The custom-built quadrotor UAV platform used for real-world experiments. The onboard sensor suite includes a solid-state LiDAR with an IMU. An edge-grade x86 Intel-N305 processor runs the AWARE decision loop in real time.}
\label{fig:uav_platform}
\end{figure}

\begin{figure*}[]
\centering
\includegraphics[width=0.98\textwidth]{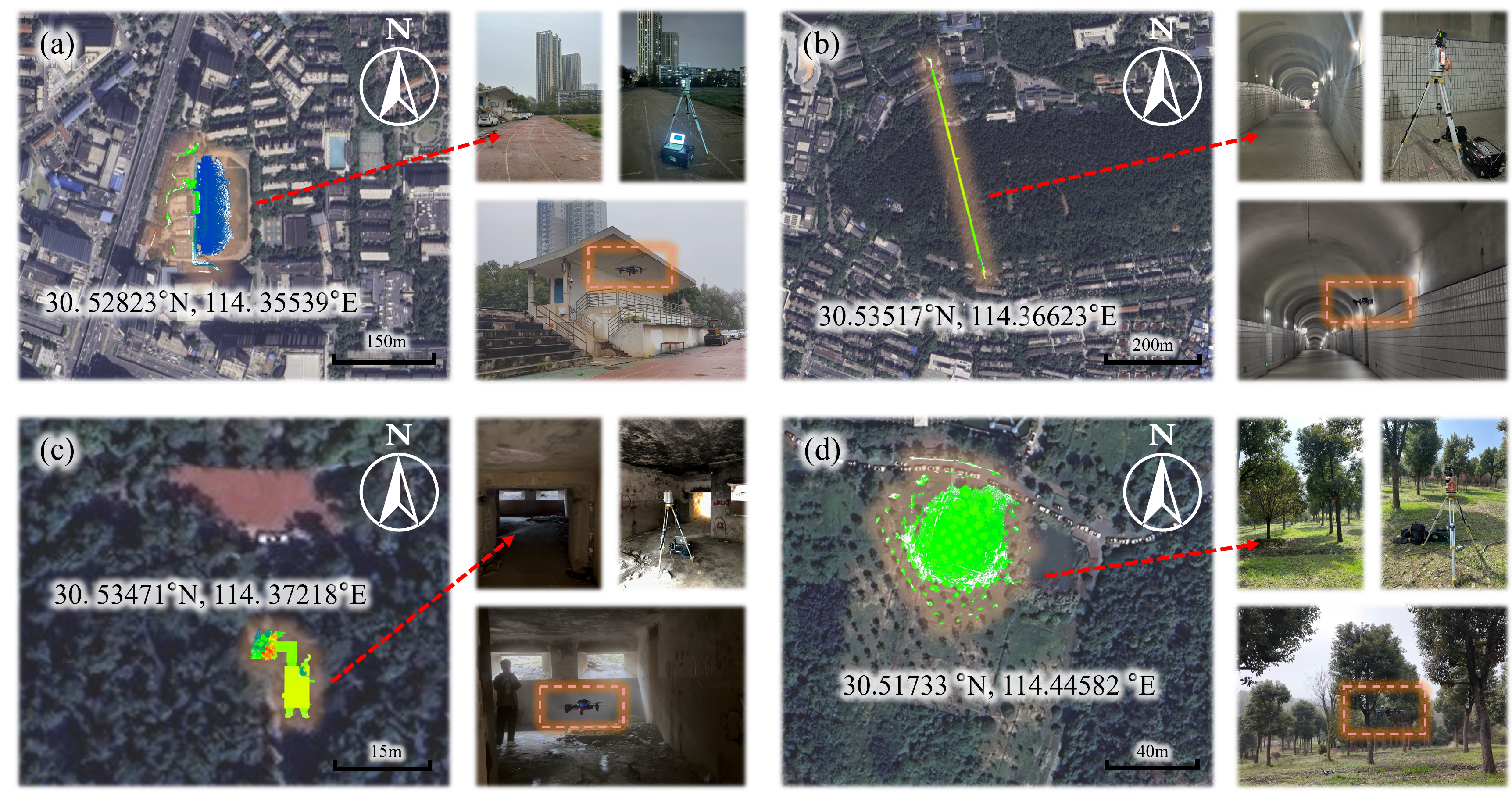}
\caption{Real-world deployment sites and representative field data acquisition. For each environment, the satellite image serves as the basemap, overlaid with the corresponding top-view point-cloud map. The accompanying photographs illustrate representative scene characteristics, TLS-based ground-truth data collection, and UAV flight during deployment. (a) Abandoned Athletic Infrastructure; (b) Underground Tunnel; (c) Abandoned Bunker; (d) Dense Forest.}
\label{fig:real_world_scenes}
\end{figure*}

To validate the practical deployability of the AWARE framework, we conducted field experiments in a variety of challenging indoor and outdoor environments. As shown in Fig.~\ref{fig:real_world_scenes}, AWARE was deployed on a custom-built quadrotor UAV (Fig.~\ref{fig:uav_platform}) across four test sites, including an underground tunnel, an abandoned athletic infrastructure, a derelict bunker, and a dense forest, collectively spanning subterranean passages, post-disaster-like man-made facilities, and unstructured vegetation terrain. Among them, the \textit{Underground Tunnel} is a roughly 400\,m-long passage with branching corridors whose repetitive cylindrical walls induce strong axial degeneracy; the \textit{Abandoned Bunker} is a confined, dust-laden structure cluttered with debris that produces frequent occlusions and noisy returns; the \textit{Abandoned Athletic Infrastructure} is an open-air facility where intermittent feature gaps arise from the flat ground and open sky; and the \textit{Dense Forest} presents irregularly distributed tree trunks under a high canopy with few planar surfaces. 
To establish a reliable ground-truth reference, we employed a survey-grade Riegl VZ-400 terrestrial laser scanner to acquire multi-station scans at each site before the flights. The individual TLS scans were then registered into a unified, high-resolution prior map using the robust point cloud registration method BSC~\cite{dong2020registration}. Each onboard LiDAR scan was subsequently registered against the corresponding prior map via scan-to-map alignment, yielding accurate per-frame 6-DoF poses as the ground-truth benchmark. Table~\ref{tab:real_world_info} summarizes the operational statistics of each sequence.

\begin{table*}[htbp]
  \centering
  \caption{Operational statistics of the real-world flight sequences.}
  \label{tab:real_world_info}
  \renewcommand{\arraystretch}{1.1}
  \begin{tabular*}{\linewidth}{@{\extracolsep{\fill}} llccc @{}}
    \toprule
    \textbf{Scene} & \textbf{Geometric Challenges} & \textbf{Duration (s)} & \textbf{Length (m)} & \textbf{Avg. Vel. (m/s)} \\
    \midrule
    Abandoned Athletic Infrastructure & Open-air, intermittent features & 568.70 & 291.67 & 0.51 \\
    Underground Tunnel                & Axial degeneracy, branching corridors & 607.90 & 407.17 & 0.67 \\
    Abandoned Bunker                  & Confined, dusted, heavy occlusion & 147.41 & 49.05 & 0.33 \\
    Dense Forest                      & Unstructured vegetation, sparse primitives & 265.82 & 110.39 & 0.42 \\
    \midrule
    \textbf{Total}                    & -- & 1589.89 & 858.28 & 0.54 \\
    \bottomrule
  \end{tabular*}
\end{table*}

\begin{figure*}[]
\centering
\includegraphics[width=0.98\textwidth]{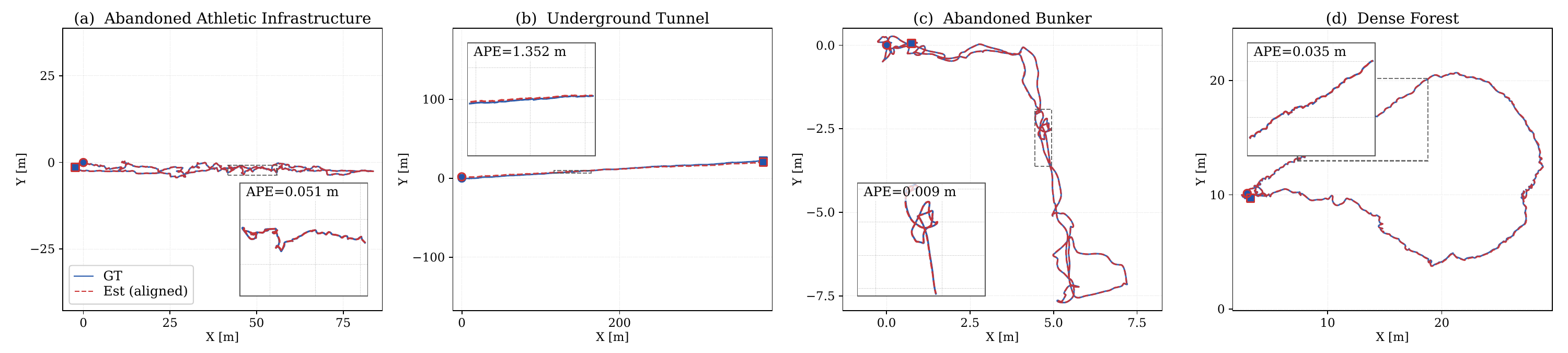}
\caption{Estimated trajectories aligned with the ground truth and the corresponding Absolute Pose Error (APE) across four real-world environments. (a)~Abandoned Athletic Infrastructure; (b)~Underground Tunnel; (c)~Abandoned Bunker; (d)~Dense Forest. Blue solid lines denote the ground truth, and red dashed lines denote the aligned estimates. Insets highlight zoomed-in views of local alignment details.}
\label{fig:real_world_traj}
\end{figure*}

Fig.~\ref{fig:real_world_traj} presents the estimated trajectories aligned with the ground truth, and Table~\ref{tab:real_world_ape} reports the corresponding APE statistics together with the drift rate, defined as the ratio between the maximum APE and the trajectory length. Beyond raw APE, the drift rate is particularly informative for real-world deployment because it normalizes the worst-case localization error by the traveled distance and therefore enables a fairer comparison across sequences of different scales. Overall, AWARE maintains bounded localization error across all four sites without any fine-tuning or domain adaptation, indicating that the policy learned in simulation transfers effectively to previously unseen physical environments.

\begin{table*}[htbp]
  \centering
  \caption{Absolute Pose Error (APE, in meters) of AWARE across real-world environments. The drift rate is computed as the ratio of APE Max to the trajectory length and serves as a normalized measure of accumulated drift over the traveled distance.}
  \label{tab:real_world_ape}
  \renewcommand{\arraystretch}{1.1}
  \begin{tabular*}{\linewidth}{@{\extracolsep{\fill}} lcccc @{}}
    \toprule
    \textbf{Scene} & \textbf{APE Mean (m)} & \textbf{APE RMSE (m)} & \textbf{APE Max (m)} & \textbf{Drift Rate (\%)} \\
    \midrule
    Abandoned Athletic Infrastructure & 0.0441 & 0.0510 & 0.1589 & 0.0545 \\
    Underground Tunnel                & 1.2008 & 1.3516 & 2.6868 & 0.6598 \\
    Abandoned Bunker                  & 0.0081 & 0.0091 & 0.0281 & 0.0573 \\
    Dense Forest                      & 0.0333 & 0.0355 & 0.0859 & 0.0778 \\
    \midrule
    \textbf{Average}                   & 0.3216 & 0.3618 & 0.7399 & 0.2124 \\
    \bottomrule
  \end{tabular*}
\end{table*}

The scene-wise results provide further insight into how the learned policy responds to different geometric conditions. The \textit{Abandoned Athletic Infrastructure} and the \textit{Dense Forest} both yield low error levels, with RMSE values of 0.0510\,m and 0.0355\,m and drift rates of 0.0545\% and 0.0778\%, respectively. Although these two environments differ substantially in appearance, both contain spatially distributed vertical structures, such as steel frames and tree trunks, that offer persistent lateral constraints for scan registration. This behavior is consistent with the strong performance observed in the simulated \textit{Building} and \textit{Forest} scenes in Table~\ref{tab:ape_results}, suggesting that the policy generalizes well when the real-world geometry matches the structural cues encountered during training.

The \textit{Abandoned Bunker} achieves the lowest error among all real-world sites, with an APE mean of 0.0081\,m, an RMSE of 0.0091\,m, and a maximum error of only 0.0281\,m. This result is notable because the bunker introduces disturbances that are only weakly represented in simulation, including airborne dust, partial occlusions, and cluttered debris. The consistently small error indicates that the panoramic depth representation and the resulting control policy remain stable even when individual measurements are degraded by noise and temporary visibility loss. From an application perspective, this robustness is particularly relevant for inspection and search missions in confined, degraded environments.

The most challenging environment is the \textit{Underground Tunnel}, where the mean APE rises to 1.2008\,m and the drift rate reaches 0.6598\%. This degradation is expected because the long corridor, repetitive cylindrical walls, and limited lateral structure create a strongly degenerate sensing geometry. Importantly, however, the error remains bounded over a 407.17\,m trajectory, and Fig.~\ref{fig:real_world_traj}(b) shows that the estimated path still follows the global trend of the reference trajectory. This observation is consistent with the simulation results in tunnel-like scenes, where AWARE also outperformed fixed scanning strategies under severe observability loss. The field evidence therefore suggests that the policy has learned a geometry-aware adaptation mechanism rather than a scene-specific control heuristic.

Taken together, the real-world experiments support two main conclusions. First, a policy trained entirely in simulation can be deployed on a physical UAV without additional adaptation while maintaining reliable localization accuracy across structurally diverse environments. Second, the consistently low drift in open facilities, cluttered indoor structures, and unstructured forests, together with the bounded error in the tunnel sequence, indicates that the RL-guided weight adjustment responds to local observability conditions in a physically meaningful manner. These findings strengthen the claim that AWARE captures a transferable active perception strategy instead of overfitting to a narrow set of training scenes.

\subsection{Time Performance Analysis}

For field deployment on autonomous UAVs, algorithmic efficiency is just as critical as localization accuracy. To demonstrate that AWARE satisfies strict real-time constraints, we conducted latency analysis of our active control module using an edge-grade x86 Intel-N305 processor. In this setup, the framework operates in a continuous loop: digesting point cloud streams, executing the RL network for feature assessment, and generating active scanning actions for the MPC solver.

\begin{figure}[]
\centering
\includegraphics[width=0.98\linewidth]{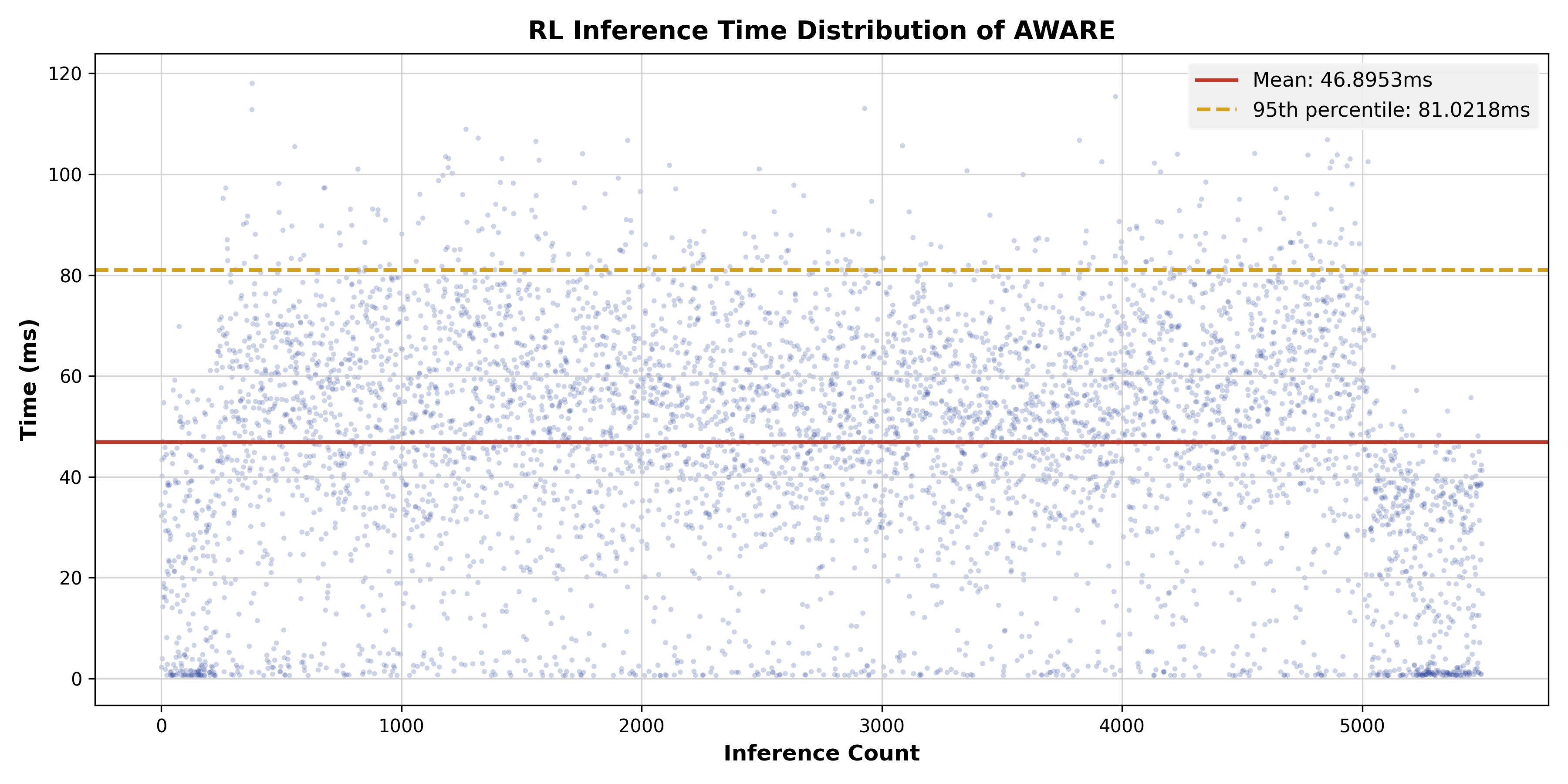}
\caption{Scatter plot and statistical distribution of the computing time required by the AWARE decision loop. The mean execution time is $46.90$~ms, with a 95th percentile limit of $81.02$~ms.}
\label{fig:rl_inference_time_distribution}
\end{figure}

The timing statistics collected during our trials are summarized in Fig.~\ref{fig:rl_inference_time_distribution}. Overall, the pipeline achieves an average processing time of $46.90$~ms, and $95\%$ of the control iterations finish within $81.02$~ms. It is worth noting that AWARE derives its control parameters dynamically based on geometric features in the environment. Since the LiDAR operates at a scanning frequency of $10$~Hz, the control module must compute its decision before the next point cloud arrives. Our timing results confirm that the algorithm's latency is bounded within this $100$~ms window. This efficiency guarantees that the active scanning policy remains tightly synchronized with real-time sensory updates. Furthermore, this compact computational footprint leaves sufficient CPU resources available for simultaneously running other essential autonomy nodes, including the primary LIO estimator, dense mapping threads, and high-level flight planners.

%% file: chapter/7Discussion.tex
\section{Discussion and future work}
\label{Discussion}

\subsection{Interpreting Yaw Exploration and Scene-Driven Observability Adapting}

The goal of this discussion is to clarify why AWARE improves localization through whole-body yaw adaptation. In our framework, yaw is not merely an attitude variable but a gimbal-like sensing action: by reorienting the onboard LiDAR, the UAV changes the visible geometry, which can improve observability when the forward-looking view becomes degenerate. This effect is especially important in corridor-like, tunnel-like, or partially occluded regions, where a fixed heading tends to overemphasize repeated or nearly parallel structures and thus yields weak lateral constraints for scan registration. The role of the hybrid RL-MPC controller is to convert this scene-dependent sensing demand into dynamically feasible yaw behavior. The RL policy modulates how strongly observability should be emphasized, while the MPC realizes that preference under tracking, smoothness, and safety constraints. Here, HITL references and SFC constraints mainly define the task and safety conditions; the localization benefit itself comes from observability-aware yaw adaptation. To examine whether this mechanism is physically meaningful, we analyze two interpretable signals collected during real-world deployment: the commanded yaw rate $\dot{\psi}$ and the RL-modulated observability weight. 

\begin{figure}[]
\centering
\includegraphics[width=0.98\linewidth]{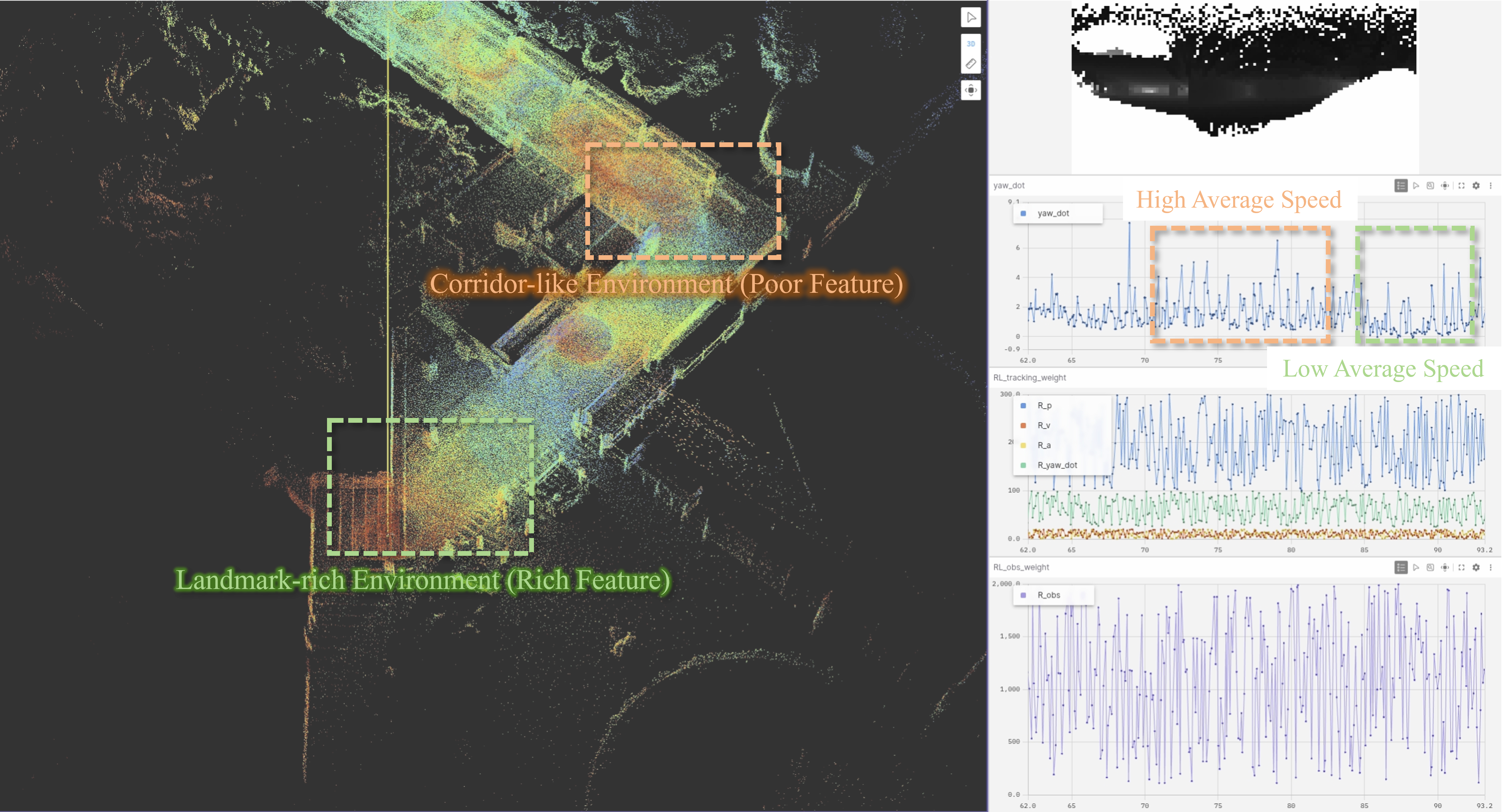}
\caption{Guided by the hybrid RL-MPC framework, AWARE adaptively modulates the commanded yaw rate $\dot{\psi}$ according to scene features. Left: representative corridor-like (feature-poor) and landmark-rich (feature-rich) subregions. Right: corresponding controller traces, where the corridor-like segment exhibits a higher average $\dot{\psi}$, whereas the landmark-rich segment requires a lower average $\dot{\psi}$.}
\label{fig:yaw_dot_analysis}
\end{figure}

Fig.~\ref{fig:yaw_dot_analysis} reveals a consistent trend: the UAV rotates faster in the corridor-like region and more slowly in the landmark-rich region. This behavior is consistent with the performance trends reported in Table~\ref{tab:ape_results} and Table~\ref{tab:real_world_ape}: the clearest benefit of AWARE appears in observability-limited environments, whereas geometrically rich scenes already require less aggressive viewpoint exploration. In weakly structured areas, maintaining a narrow forward-looking direction can leave the estimator in an unfavorable information geometry when the visible surfaces are dominated by nearly parallel walls or distant planar structures. A larger yaw rate broadens the local viewpoint search and changes both the subset of observed geometry and the corresponding incidence angles, thereby exposing stronger constraints such as wall intersections, openings, ceiling-floor boundaries, cluttered side structures, or transitions near the boundary of the current field of view. The higher $\dot{\psi}$ observed in the corridor-like segment can therefore be interpreted as targeted viewpoint reallocation under geometric degeneracy. Conversely, the lower $\dot{\psi}$ in the landmark-rich segment indicates that the controller does not rotate aggressively when sufficient multi-directional constraints are already available, which is consistent with the low localization error observed in feature-rich scenes.

The same mechanism is reflected in the RL policy's scene-dependent modulation of $\lambda_{obs}$. Table~\ref{tab:r_obs_real} summarizes the mean observability weight measured in four real-world environments. A clear ordering emerges. The \textit{Underground Tunnel} yields the highest average weight, followed closely by the \textit{Abandoned Bunker}, while the \textit{Dense Forest} yields by far the lowest value. This ordering agrees well with the geometric difficulty of each site and with the real-world error trends reported in Table~\ref{tab:real_world_ape}. The tunnel is the most degenerate case because long axial corridors and repetitive surfaces provide limited lateral constraints, so the controller assigns the highest priority to observability preservation. The bunker also requires a large weight because occlusion, clutter, and dust can locally degrade registration quality even though the surrounding structure is dense. The abandoned athletic infrastructure exhibits an intermediate value, which is reasonable for an open environment with intermittent feature gaps. In contrast, the dense forest naturally supplies abundant multi-directional constraints from tree trunks and nearby vegetation, allowing the controller to maintain reliable localization without unnecessarily amplifying the observability term.

\begin{table}[!htbp]
	\centering
	\caption{Average RL modulated observability weight $\lambda_{obs}$ across real world scenes. Larger values indicate stronger emphasis on observability in the MPC objective.}
	\label{tab:r_obs_real}
	\renewcommand{\arraystretch}{1.1}
	\begin{tabular}{lc}
		\hline
		{\bfseries Scene} & {\bfseries Mean $\mathbf{\lambda_{obs}}$} \\
		\hline
		Abandoned Athletic Infrastructure & 1124.14 \\
		Underground Tunnel & 1585.54 \\
		Abandoned Bunker & 1544.85 \\
		Dense Forest & 268.30 \\
		\hline
	\end{tabular}
\end{table}

Taken together, the yaw behavior and the weight statistics suggest that AWARE does not simply rotate more in every scene. Instead, it reallocates yaw effort to environments where observability is limited and relaxes that effort when geometry is already informative. This division of roles also improves interpretability: whole-body yaw serves as the lightweight sensing mechanism, RL decides when observability should be prioritized, and MPC determines how that preference can be executed within the operating constraints. The empirical trends in Fig.~\ref{fig:yaw_dot_analysis} and Table~\ref{tab:r_obs_real} therefore support the interpretation that the learned policy captures a scene-aware mechanism for reducing localization drift rather than a scene-specific motion heuristic.

\subsection{Future Directions}

Future work can extend AWARE in two closely related directions. One is to move from a single UAV to multiple UAVs that share compact observability cues and coordinate complementary viewpoints, which may improve coverage efficiency and localization robustness in large scale or strongly degenerate environments. The other is to integrate AWARE with immersive VR interfaces so that operators can inspect the reconstructed scene, uncertainty distribution, and suggested viewing directions more intuitively while issuing high level commands in a more natural manner. Combining cooperative active perception with richer human interaction may further broaden the applicability of AWARE in exploration, inspection, and emergency response tasks.

%% file: chapter/8Conclusion.tex
\section{Conclusion}
\label{sec:conclusion}

This paper presented AWARE, a bio-inspired whole-body active yawing framework that enhances LiDAR-inertial state estimation on resource-constrained UAVs by exploiting the platform's own rotational agility, eliminating the need for additional mechanical actuation. A unified panoramic depth representation coupled with a Fisher Information Matrix-based observability metric enables the system to evaluate geometric richness across the full yaw space and actively steer toward the viewing direction that maximizes information gain. A hybrid RL-MPC architecture, where a lightweight RL agent dynamically modulates MPC cost weights based on real-time environmental context, resolves the trade-off between active perception and flight stability without manual parameter tuning. Safe Flight Corridor constraints further decouple the operator's translational intent from autonomous yaw optimization, ensuring operational safety. Extensive simulated and real-world experiments demonstrate that AWARE consistently outperforms passive and static optimization baselines across diverse scenarios while satisfying real-time computational requirements for onboard deployment. Future work will extend AWARE toward multi-UAV coordination with complementary observability cues and toward immersive VR interfaces that support more intuitive human interaction during deployment.

%% file: reference.bib
@article{li2025ua,
  title={Ua-mpc: Uncertainty-aware model predictive control for motorized lidar odometry},
  author={Li, Jianping and Xu, Xinhang and Liu, Jinxin and Cao, Kun and Yuan, Shenghai and Xie, Lihua},
  journal={IEEE Robotics and Automation Letters},
  year={2025},
  publisher={IEEE}
}

@article{li2026aeos,
  title={Aeos: Active environment-aware optimal scanning control for uav lidar-inertial odometry in complex scenes},
  author={Li, Jianping and Xu, Xinhang and Liu, Zhongyuan and Yuan, Shenghai and Cao, Muqing and Xie, Lihua},
  journal={ISPRS Journal of Photogrammetry and Remote Sensing},
  volume={232},
  pages={476--491},
  year={2026},
  publisher={Elsevier}
}

@article{xu2022fast,
  title={Fast-lio2: Fast direct lidar-inertial odometry},
  author={Xu, Wei and Cai, Yixi and He, Dongjiao and Lin, Jiarong and Zhang, Fu},
  journal={IEEE Transactions on Robotics},
  volume={38},
  number={4},
  pages={2053--2073},
  year={2022},
  publisher={IEEE}
}

@article{liu2025slope,
  title={Slope inspection under dense vegetation using LiDAR-based quadrotors},
  author={Liu, Wenyi and Ren, Yunfan and Guo, Rui and Kong, Vickie WW and Hung, Anthony SP and Zhu, Fangcheng and Cai, Yixi and Wu, Huajie and Zou, Yuying and Zhang, Fu},
  journal={Nature Communications},
  volume={16},
  number={1},
  pages={7411},
  year={2025},
  publisher={Nature Publishing Group UK London}
}

@article{zhang2025armor,
  title={ARMOR: Adaptive Meshing with Reinforcement Optimization for Real-time 3D Monitoring in Unexposed Scenes},
  author={Zhang, Yizhe and Li, Jianping and Zhao, Xin and Liang, Fuxun and Dong, Zhen and Yang, Bisheng},
  journal={arXiv preprint arXiv:2504.19624},
  year={2025}
}

@inproceedings{messikommer2024reinforcement,
  title={Reinforcement learning meets visual odometry},
  author={Messikommer, Nico and Cioffi, Giovanni and Gehrig, Mathias and Scaramuzza, Davide},
  booktitle={European Conference on Computer Vision},
  pages={76--92},
  year={2024},
  organization={Springer}
}

@inproceedings{xu2025flying,
  title={Flying on point clouds with reinforcement learning},
  author={Xu, Guangtong and Wu, Tianyue and Wang, Zihan and Wang, Qianhao and Gao, Fei},
  booktitle={2025 IEEE/RSJ International Conference on Intelligent Robots and Systems (IROS)},
  pages={7231--7238},
  year={2025},
  organization={IEEE}
}

@article{zhang2026high,
  title={High-Speed Vision-Based Flight in Clutter with Safety-Shielded Reinforcement Learning},
  author={Zhang, Jiarui and Lei, Chengyong and Dai, Chengjiang and Wang, Lijie and Han, Zhichao and Gao, Fei},
  journal={arXiv preprint arXiv:2602.08653},
  year={2026}
}

@inproceedings{romero2024actor,
  title={Actor-critic model predictive control},
  author={Romero, Angel and Song, Yunlong and Scaramuzza, Davide},
  booktitle={2024 IEEE International Conference on Robotics and Automation (ICRA)},
  pages={14777--14784},
  year={2024},
  organization={IEEE}
}

@article{romero2025actor,
  title={Actor--Critic Model Predictive Control: Differentiable Optimization Meets Reinforcement Learning for Agile Flight},
  author={Romero, Angel and Aljalbout, Elie and Song, Yunlong and Scaramuzza, Davide},
  journal={IEEE Transactions on Robotics},
  volume={42},
  pages={673--692},
  year={2025},
  publisher={IEEE}
}

@article{zhu2025flare,
  title={FLARE: Fast Autonomous Aerial Exploration in Large-Scale 3D Scenarios Using Actively Rotated LiDAR},
  author={Zhu, Zhiwen and Fang, Yuhao and Xiao, Xulin and Lyu, Ximin and Mei, Jie and Zhou, Boyu},
  journal={IEEE Transactions on Automation Science and Engineering},
  volume={22},
  pages={24077--24091},
  year={2025},
  publisher={IEEE}
}

@article{alismail2015automatic,
  title={Automatic calibration of spinning actuated lidar internal parameters},
  author={Alismail, Hatem and Browning, Brett},
  journal={Journal of Field Robotics},
  volume={32},
  number={5},
  pages={723--747},
  year={2015},
  publisher={Wiley Online Library}
}

@inproceedings{zhang2014loam,
  title={LOAM: Lidar odometry and mapping in real-time.},
  author={Zhang, Ji and Singh, Sanjiv and others},
  booktitle={Robotics: Science and systems},
  volume={2},
  number={9},
  pages={1--9},
  year={2014},
  organization={Berkeley, CA}
}

@inproceedings{zhen2017robust,
  title={Robust localization and localizability estimation with a rotating laser scanner},
  author={Zhen, Weikun and Zeng, Sam and Soberer, Sebastian},
  booktitle={2017 IEEE international conference on robotics and automation (ICRA)},
  pages={6240--6245},
  year={2017},
  organization={IEEE}
}

@article{gong2023rss,
  title={RSS-LIWOM: Rotating solid-state LiDAR for robust LiDAR-Inertial-Wheel odometry and mapping},
  author={Gong, Shunjie and Shi, Chenghao and Zhang, Hui and Lu, Huimin and Zeng, Zhiwen and Chen, Xieyuanli},
  journal={Remote Sensing},
  volume={15},
  number={16},
  pages={4040},
  year={2023},
  publisher={MDPI}
}

@Article{mi11050456,
AUTHOR = {Wang, Dingkang and Watkins, Connor and Xie, Huikai},
TITLE = {MEMS Mirrors for LiDAR: A Review},
JOURNAL = {Micromachines},
VOLUME = {11},
YEAR = {2020},
NUMBER = {5},
ARTICLE-NUMBER = {456},
URL = {https://www.mdpi.com/2072-666X/11/5/456},
PubMedID = {32349453},
ISSN = {2072-666X},
DOI = {10.3390/mi11050456}
}

@article{juliano2022metasurface,
  title={Metasurface-enhanced light detection and ranging technology},
  author={Juliano Martins, Renato and Marinov, Emil and Youssef, M Aziz Ben and Kyrou, Christina and Joubert, Mathilde and Colmagro, Constance and G{\^a}t{\'e}, Valentin and Turbil, Colette and Coulon, Pierre-Marie and Turover, Daniel and others},
  journal={Nature communications},
  volume={13},
  number={1},
  pages={5724},
  year={2022},
  publisher={Nature Publishing Group UK London}
}

@article{li2025limo,
  title={Limo-calib: On-site fast lidar-motor calibration for quadruped robot-based panoramic 3d sensing system},
  author={Li, Jianping and Liu, Zhongyuan and Xu, Xinhang and Liu, Jinxin and Yuan, Shenghai and Xu, Fang and Xie, Lihua},
  journal={arXiv preprint arXiv:2502.12655},
  year={2025}
}

@misc{livox2024scanner,
  author       = {{Livox-SDK Team}},
  title        = {Livox scanner: Motorized LiDAR scanning software development kit},
  year         = {2024},
  howpublished = {\url{https://github.com/Livox-SDK/livox_scanner}},
  note         = {Accessed: 2026-1-13}
}

@article{chen2024design,
  title={Design of an Adaptive Lightweight Lidar to decouple robot--camera geometry},
  author={Chen, Yuyang and Wang, Dingkang and Thomas, Lenworth and Dantu, Karthik and Koppal, Sanjeev J},
  journal={IEEE Transactions on Robotics},
  volume={40},
  pages={2254--2271},
  year={2024},
  publisher={IEEE}
}

@article{chen2023self,
  title={A self-rotating, single-actuated UAV with extended sensor field of view for autonomous navigation},
  author={Chen, Nan and Kong, Fanze and Xu, Wei and Cai, Yixi and Li, Haotian and He, Dongjiao and Qin, Youming and Zhang, Fu},
  journal={Science Robotics},
  volume={8},
  number={76},
  pages={eade4538},
  year={2023},
  publisher={American Association for the Advancement of Science}
}

@inproceedings{cao2025learning,
  title={Learning dynamic weight adjustment for spatial-temporal trajectory planning in crowd navigation},
  author={Cao, Muqing and Xu, Xinhang and Yang, Yizhuo and Li, Jianping and Jin, Tongxing and Wang, Pengfei and Hung, Tzu-Yi and Lin, Guosheng and Xie, Lihua},
  booktitle={2025 IEEE International Conference on Robotics and Automation (ICRA)},
  pages={8196--8202},
  year={2025},
  organization={IEEE}
}

@article{ling2023efficacy,
  title={On the efficacy of 3d point cloud reinforcement learning},
  author={Ling, Zhan and Yao, Yunchao and Li, Xuanlin and Su, Hao},
  journal={arXiv preprint arXiv:2306.06799},
  year={2023}
}

@article{xu2025navrl,
  title={Navrl: Learning safe flight in dynamic environments},
  author={Xu, Zhefan and Han, Xinming and Shen, Haoyu and Jin, Hanyu and Shimada, Kenji},
  journal={IEEE Robotics and Automation Letters},
  year={2025},
  publisher={IEEE}
}

@article{foehn2021time,
  title={Time-optimal planning for quadrotor waypoint flight},
  author={Foehn, Philipp and Romero, Angel and Scaramuzza, Davide},
  journal={Science robotics},
  volume={6},
  number={56},
  pages={eabh1221},
  year={2021},
  publisher={American Association for the Advancement of Science}
}

@article{romero2022model,
  title={Model predictive contouring control for time-optimal quadrotor flight},
  author={Romero, Angel and Sun, Sihao and Foehn, Philipp and Scaramuzza, Davide},
  journal={IEEE Transactions on Robotics},
  volume={38},
  number={6},
  pages={3340--3356},
  year={2022},
  publisher={IEEE}
}

@article{romero2022time,
  title={Time-optimal online replanning for agile quadrotor flight},
  author={Romero, Angel and Penicka, Robert and Scaramuzza, Davide},
  journal={IEEE Robotics and Automation Letters},
  volume={7},
  number={3},
  pages={7730--7737},
  year={2022},
  publisher={IEEE}
}

@article{song2023reaching,
  title={Reaching the limit in autonomous racing: Optimal control versus reinforcement learning},
  author={Song, Yunlong and Romero, Angel and M{\"u}ller, Matthias and Koltun, Vladlen and Scaramuzza, Davide},
  journal={Science Robotics},
  volume={8},
  number={82},
  pages={eadg1462},
  year={2023},
  publisher={American Association for the Advancement of Science}
}

@inproceedings{xing2024contrastive,
  title={Contrastive learning for enhancing robust scene transfer in vision-based agile flight},
  author={Xing, Jiaxu and Bauersfeld, Leonard and Song, Yunlong and Xing, Chunwei and Scaramuzza, Davide},
  booktitle={2024 IEEE International Conference on Robotics and Automation (ICRA)},
  pages={5330--5337},
  year={2024},
  organization={IEEE}
}

@article{wang2022geometrically,
  title={Geometrically constrained trajectory optimization for multicopters},
  author={Wang, Zhepei and Zhou, Xin and Xu, Chao and Gao, Fei},
  journal={IEEE Transactions on Robotics},
  volume={38},
  number={5},
  pages={3259--3278},
  year={2022},
  publisher={IEEE}
}

@inproceedings{ren2022bubble,
  title={Bubble planner: Planning high-speed smooth quadrotor trajectories using receding corridors},
  author={Ren, Yunfan and Zhu, Fangcheng and Liu, Wenyi and Wang, Zhepei and Lin, Yi and Gao, Fei and Zhang, Fu},
  booktitle={2022 IEEE/RSJ International Conference on Intelligent Robots and Systems (IROS)},
  pages={6332--6339},
  year={2022},
  organization={IEEE}
}

@article{liu2017planning,
  title={Planning dynamically feasible trajectories for quadrotors using safe flight corridors in 3-d complex environments},
  author={Liu, Sikang and Watterson, Michael and Mohta, Kartik and Sun, Ke and Bhattacharya, Subhrajit and Taylor, Camillo J and Kumar, Vijay},
  journal={IEEE Robotics and Automation Letters},
  volume={2},
  number={3},
  pages={1688--1695},
  year={2017},
  publisher={IEEE}
}

@article{zhou2020ego,
  title={Ego-planner: An esdf-free gradient-based local planner for quadrotors},
  author={Zhou, Xin and Wang, Zhepei and Ye, Hongkai and Xu, Chao and Gao, Fei},
  journal={IEEE Robotics and Automation Letters},
  volume={6},
  number={2},
  pages={478--485},
  year={2020},
  publisher={IEEE}
}

@article{zhou2019robust,
  title={Robust and efficient quadrotor trajectory generation for fast autonomous flight},
  author={Zhou, Boyu and Gao, Fei and Wang, Luqi and Liu, Chuhao and Shen, Shaojie},
  journal={IEEE Robotics and Automation Letters},
  volume={4},
  number={4},
  pages={3529--3536},
  year={2019},
  publisher={IEEE}
}

@inproceedings{mellinger2011minimum,
  title={Minimum snap trajectory generation and control for quadrotors},
  author={Mellinger, Daniel and Kumar, Vijay},
  booktitle={2011 IEEE international conference on robotics and automation},
  pages={2520--2525},
  year={2011},
  organization={IEEE}
}

@article{aucone2024synergistic,
  title={Synergistic morphology and feedback control for traversal of unknown compliant obstacles with aerial robots},
  author={Aucone, Emanuele and Geckeler, Christian and Morra, Daniele and Pallottino, Lucia and Mintchev, Stefano},
  journal={Nature Communications},
  volume={15},
  number={1},
  pages={2646},
  year={2024},
  publisher={Nature Publishing Group UK London}
}

@article{liu2023integrated,
  title={Integrated planning and control for quadrotor navigation in presence of suddenly appearing objects and disturbances},
  author={Liu, Wenyi and Ren, Yunfan and Zhang, Fu},
  journal={IEEE Robotics and Automation Letters},
  volume={9},
  number={1},
  pages={899--906},
  year={2023},
  publisher={IEEE}
}

@article{kong2021avoiding,
  title={Avoiding dynamic small obstacles with onboard sensing and computation on aerial robots},
  author={Kong, Fanze and Xu, Wei and Cai, Yixi and Zhang, Fu},
  journal={IEEE Robotics and Automation Letters},
  volume={6},
  number={4},
  pages={7869--7876},
  year={2021},
  publisher={IEEE}
}

@article{ebadi2023present,
  title={Present and future of slam in extreme environments: The darpa subt challenge},
  author={Ebadi, Kamak and Bernreiter, Lukas and Biggie, Harel and Catt, Gavin and Chang, Yun and Chatterjee, Arghya and Denniston, Christopher E and Desch{\^e}nes, Simon-Pierre and Harlow, Kyle and Khattak, Shehryar and others},
  journal={IEEE Transactions on Robotics},
  volume={40},
  pages={936--959},
  year={2023},
  publisher={IEEE}
}

@article{yin2024survey,
  title={A survey on global lidar localization: Challenges, advances and open problems},
  author={Yin, Huan and Xu, Xuecheng and Lu, Sha and Chen, Xieyuanli and Xiong, Rong and Shen, Shaojie and Stachniss, Cyrill and Wang, Yue},
  journal={International Journal of Computer Vision},
  volume={132},
  number={8},
  pages={3139--3171},
  year={2024},
  publisher={Springer}
}

@inproceedings{zhang2024lio,
  title={As-lio: Spatial overlap guided adaptive sliding window lidar-inertial odometry for aggressive fov variation},
  author={Zhang, Tianxiang and Zhang, Xuanxuan and Liao, Zongbo and Xia, Xin and Li, You},
  booktitle={2024 IEEE/RSJ International Conference on Intelligent Robots and Systems (IROS)},
  pages={10829--10836},
  year={2024},
  organization={IEEE}
}

@article{kaul2016continuous,
  title={Continuous-time three-dimensional mapping for micro aerial vehicles with a passively actuated rotating laser scanner},
  author={Kaul, Lukas and Zlot, Robert and Bosse, Michael},
  journal={Journal of Field Robotics},
  volume={33},
  number={1},
  pages={103--132},
  year={2016},
  publisher={Wiley Online Library}
}

@inproceedings{jameson2012lockheed,
  title={Lockheed martin's samarai nano air vehicle: Challenges, research, and realization},
  author={Jameson, Steve and Fregene, Kingsley and Chang, Ming and Allen, Ned and Youngren, Harold and Scroggins, Joe},
  booktitle={50th AIAA aerospace sciences meeting including the new horizons forum and aerospace exposition},
  pages={584},
  year={2012}
}

@inproceedings{bartolomei2021semantic,
  title={Semantic-aware active perception for uavs using deep reinforcement learning},
  author={Bartolomei, Luca and Teixeira, Lucas and Chli, Margarita},
  booktitle={2021 IEEE/RSJ International Conference on Intelligent Robots and Systems (IROS)},
  pages={3101--3108},
  year={2021},
  organization={IEEE}
}

@inproceedings{cui2014autonomous,
  title={Autonomous navigation of UAV in forest},
  author={Cui, Jin Qiang and Lai, Shupeng and Dong, Xiangxu and Liu, Peidong and Chen, Ben M and Lee, Tong H},
  booktitle={2014 International Conference on Unmanned Aircraft Systems (ICUAS)},
  pages={726--733},
  year={2014},
  organization={IEEE}
}

@article{greenwood2019applications,
  title={Applications of UAVs in civil infrastructure},
  author={Greenwood, William W and Lynch, Jerome P and Zekkos, Dimitrios},
  journal={Journal of infrastructure systems},
  volume={25},
  number={2},
  pages={04019002},
  year={2019},
  publisher={American Society of Civil Engineers}
}

@article{liu2025artemis,
  title={ARTEMIS: A real-time efficient ortho-mapping and thematic identification system for UAV-based rapid response},
  author={Liu, Yijun and Akbar, Akram and Yu, Ting and Yu, Yunlong and Kong, Yuanhang and Gao, Jingwen and Wang, Honghao and Li, Yanyi and Zhao, Hongduo and Liu, Chun},
  journal={ISPRS Journal of Photogrammetry and Remote Sensing},
  volume={229},
  pages={396--421},
  year={2025},
  publisher={Elsevier}
}

@article{drescher2010fidelity,
  title={Fidelity of adaptive phototaxis},
  author={Drescher, Knut and Goldstein, Raymond E and Tuval, Idan},
  journal={Proceedings of the National Academy of Sciences},
  volume={107},
  number={25},
  pages={11171--11176},
  year={2010},
  publisher={National Academy of Sciences}
}

@article{li2023whu,
  title={WHU-helmet: A helmet-based multisensor SLAM dataset for the evaluation of real-time 3-D mapping in large-scale GNSS-denied environments},
  author={Li, Jianping and Wu, Weitong and Yang, Bisheng and Zou, Xianghong and Yang, Yandi and Zhao, Xin and Dong, Zhen},
  journal={IEEE Transactions on Geoscience and Remote Sensing},
  volume={61},
  pages={1--16},
  year={2023},
  publisher={IEEE}
}

@article{petravcek2021large,
  title={Large-scale exploration of cave environments by unmanned aerial vehicles},
  author={Petr{\'a}{\v{c}}ek, Pavel and Kr{\'a}tk{\`y}, V{\'\i}t and Petrl{\'\i}k, Mat{\v{e}}j and B{\'a}{\v{c}}a, Tom{\'a}{\v{s}} and Kratochv{\'\i}l, Radim and Saska, Martin},
  journal={IEEE Robotics and Automation Letters},
  volume={6},
  number={4},
  pages={7596--7603},
  year={2021},
  publisher={IEEE}
}

@misc{autowarefoundation,
  author = {Autoware},
  url ={https://github.com/autowarefoundation/autoware},
  year = {2022}, 
  note = {Accessed: 2026-1-21}
}

@article{dong2020registration,
  title={Registration of large-scale terrestrial laser scanner point clouds: A review and benchmark},
  author={Dong, Zhen and Liang, Fuxun and Yang, Bisheng and Xu, Yusheng and Zang, Yufu and Li, Jianping and Wang, Yuan and Dai, Wenxia and Fan, Hongchao and Hyypp{\"a}, Juha and others},
  journal={ISPRS Journal of Photogrammetry and Remote Sensing},
  volume={163},
  pages={327--342},
  year={2020},
  publisher={Elsevier}
}

@article{cadena2017past,
  title={Past, present, and future of simultaneous localization and mapping: Toward the robust-perception age},
  author={Cadena, Cesar and Carlone, Luca and Carrillo, Henry and Latif, Yasir and Scaramuzza, Davide and Neira, Jos{\'e} and Reid, Ian and Leonard, John J},
  journal={IEEE Transactions on robotics},
  volume={32},
  number={6},
  pages={1309--1332},
  year={2017},
  publisher={IEEE}
}

@article{li20193d,
  title={3D forest mapping using a low-cost UAV laser scanning system: Investigation and comparison},
  author={Li, Jianping and Yang, Bisheng and Cong, Yangzi and Cao, Lin and Fu, Xiaoyao and Dong, Zhen},
  journal={Remote Sensing},
  volume={11},
  number={6},
  pages={717},
  year={2019},
  publisher={MDPI}
}

@inproceedings{lin2020loam,
  title={Loam livox: A fast, robust, high-precision LiDAR odometry and mapping package for LiDARs of small FoV},
  author={Lin, Jiarong and Zhang, Fu},
  booktitle={2020 IEEE International Conference on Robotics and Automation (ICRA)},
  pages={3126--3131},
  year={2020},
  organization={IEEE}
}

@article{park2020applications,
  title={Applications of unmanned aerial vehicles in mining from exploration to reclamation: A review},
  author={Park, Sebeom and Choi, Yosoon},
  journal={Minerals},
  volume={10},
  number={8},
  pages={663},
  year={2020},
  publisher={MDPI}
}

@article{hao2026mapping,
  title={Mapping aboveground tree biomass and uncertainty using an upscaling approach: A case study of the larch forests in northeastern China using UAV laser scanning data},
  author={Hao, Yuanshuo and Pukkala, Timo and Liu, Xin and Quan, Ying and Dong, Lihu and Li, Fengri},
  journal={ISPRS Journal of Photogrammetry and Remote Sensing},
  volume={231},
  pages={595--607},
  year={2026},
  publisher={Elsevier}
}

@article{puniach2021application,
  title={Application of UAV-based orthomosaics for determination of horizontal displacement caused by underground mining},
  author={Puniach, Edyta and Gruszczy{\'n}ski, Wojciech and {\'C}wi{\k{a}}ka{\l}a, Pawe{\l} and Matwij, Wojciech},
  journal={ISPRS Journal of Photogrammetry and Remote Sensing},
  volume={174},
  pages={282--303},
  year={2021},
  publisher={Elsevier}
}

@article{khan2022emerging,
  title={Emerging UAV technology for disaster detection, mitigation, response, and preparedness},
  author={Khan, Amina and Gupta, Sumeet and Gupta, Sachin Kumar},
  journal={Journal of Field Robotics},
  volume={39},
  number={6},
  pages={905--955},
  year={2022},
  publisher={Wiley Online Library}
}

@article{zhang2025faem,
  title={FAEM: Fast autonomous exploration for UAV in large-scale unknown environments using LiDAR-based mapping},
  author={Zhang, Xu and Wang, Jiqiang and Wang, Shuwen and Wang, Mengfei and Wang, Tao and Feng, Zhuowen and Zhu, Shibo and Zheng, Enhui},
  journal={Drones},
  volume={9},
  number={6},
  pages={423},
  year={2025},
  publisher={MDPI}
}

@article{zhang2025lidar,
  title={Lidar-based autonomous exploration method of mobile robot using deep reinforcement learning in unknown environments},
  author={Zhang, Chizhou and Chen, Mingsong and Lin, Yongcheng and Cheng, Huaguo and Wang, Guanqiang and Li, Kai and Tang, Jun and Li, Zehao and Shen, Lirui and Wang, Qiu},
  journal={IEEE Transactions on Instrumentation and Measurement},
  year={2025},
  publisher={IEEE}
}

@article{ren2025survey,
  title={A Survey on LiDAR-based autonomous aerial vehicles},
  author={Ren, Yunfan and Cai, Yixi and Li, Haotian and Chen, Nan and Zhu, Fangcheng and Yin, Longji and Kong, Fanze and Li, Rundong and Zhang, Fu},
  journal={IEEE/ASME Transactions on Mechatronics},
  year={2025},
  publisher={IEEE}
}

@article{zhang2020falco,
  title={Falco: Fast likelihood-based collision avoidance with extension to human-guided navigation},
  author={Zhang, Ji and Hu, Chen and Chadha, Rushat Gupta and Singh, Sanjiv},
  journal={Journal of Field Robotics},
  volume={37},
  number={8},
  pages={1300--1313},
  year={2020},
  publisher={Wiley Online Library}
}

@article{cong2025adaptive,
  title={Adaptive Covariance Matrix for UAV-Based Visual--Inertial Navigation Systems Using Gaussian Formulas},
  author={Cong, Yangzi and Su, Wenbin and Jiang, Nan and Zong, Wenpeng and Li, Long and Xu, Yan and Xu, Tianhe and Wu, Paipai},
  journal={Sensors},
  volume={25},
  number={15},
  pages={4745},
  year={2025},
  publisher={MDPI}
}

@inproceedings{zhao2021super,
  title={Super odometry: IMU-centric LiDAR-visual-inertial estimator for challenging environments},
  author={Zhao, Shibo and Zhang, Hengrui and Wang, Peng and Nogueira, Lucas and Scherer, Sebastian},
  booktitle={2021 IEEE/RSJ International Conference on Intelligent Robots and Systems (IROS)},
  pages={8729--8736},
  year={2021},
  organization={IEEE}
}

@article{wang2021lightweight,
  title={Lightweight 3-D localization and mapping for solid-state LiDAR},
  author={Wang, Han and Wang, Chen and Xie, Lihua},
  journal={IEEE Robotics and Automation Letters},
  volume={6},
  number={2},
  pages={1801--1807},
  year={2021},
  publisher={IEEE}
}

@inproceedings{oleynikova2017voxblox,
  title={Voxblox: Incremental 3D Euclidean signed distance fields for on-board MAV planning},
  author={Oleynikova, Helen and Taylor, Zachary and Fehr, Marius and Siegwart, Roland and Nieto, Juan},
  booktitle={2017 IEEE/RSJ International Conference on Intelligent Robots and Systems (IROS)},
  pages={1366--1373},
  year={2017},
  organization={IEEE}
}

@inproceedings{han2019fiesta,
  title={FIESTA: Fast incremental euclidean distance fields for online motion planning of aerial robots},
  author={Han, Luxin and Gao, Fei and Zhou, Boyu and Shen, Shaojie},
  booktitle={2019 IEEE/RSJ International Conference on Intelligent Robots and Systems (IROS)},
  pages={4423--4430},
  year={2019},
  organization={IEEE}
}

@inproceedings{falanga2018pampc,
  title={PAMPC: Perception-aware model predictive control for quadrotors},
  author={Falanga, Davide and Foehn, Philipp and Lu, Peng and Scaramuzza, Davide},
  booktitle={2018 IEEE/RSJ International Conference on Intelligent Robots and Systems (IROS)},
  pages={1--8},
  year={2018},
  organization={IEEE}
}
